\documentclass[pdflatex,sn-mathphys-num]{sn-jnl}%
\usepackage{graphicx}%
\usepackage{multirow}%
\usepackage{amsmath,amssymb,amsfonts}%
\usepackage{amsthm}%
\usepackage{mathrsfs}%
\usepackage[title]{appendix}%
\usepackage{xcolor}%
\usepackage{textcomp}%
\usepackage{manyfoot}%
\usepackage{booktabs}%
\usepackage{algpseudocode}%
\usepackage{listings}%
\usepackage{siunitx}
\theoremstyle{thmstyleone}%
\usepackage[ruled,vlined,linesnumbered]{algorithm2e}
\theoremstyle{thmstyletwo}%
\usepackage{adjustbox}
\theoremstyle{thmstylethree}%
\usepackage{tikz}
\usepackage{lineno}
\usepackage{pgfplots}
\pgfplotsset{compat=1.18}
\usepackage{subcaption}
\usepgfplotslibrary{fillbetween}
\usepackage{float}
\usepackage{multirow}
\usetikzlibrary{shapes.geometric, arrows}
\usepackage{circuitikz}
\usepackage{booktabs}
\usepackage[inkscapelatex=false]{svg}
\usepackage{lineno}
\usepackage{booktabs}
\usepackage{siunitx}
\usepackage{makecell} 
\usepackage{dsfont}
\usepackage{array}
\usepackage{nameref}
\begin{document}


\title[Article Title]{pADAM: a plug-and-play all-in-one diffusion architecture for multi-physics learning}


\author[1]{\fnm{Amirhossein} \sur{Mollaali}}\email{amollaal@purdue.edu}

\author[1]{\fnm{Bongseok} \sur{Kim}}\email{kim4853@purdue.edu}

\author[2]{\fnm{Christian} \sur{Moya}}\email{cmoyacal@purdue.edu}

\author*[1,2]{\fnm{Guang} \sur{Lin}}\email{guanglin@purdue.edu}

\affil*[1]{\orgdiv{School of Mechanical Engineering}, \orgname{Purdue University}, \orgaddress{\city{West Lafayette}, \postcode{47906}, \state{IN}, \country{USA}}}

\affil*[2]{\orgdiv{Department of Mathematics}, \orgname{Purdue University}, \orgaddress{\city{West Lafayette}, \postcode{47906}, \state{IN}, \country{USA}}}


\abstract{
Generalizing across disparate physical laws remains a fundamental challenge for artificial intelligence in science. Existing deep-learning solvers are largely confined to single-equation settings, limiting transfer across physical regimes and inference tasks. Here we introduce pADAM, a unified generative framework that learns a shared probabilistic prior across heterogeneous partial differential equation families. Through a learned joint distribution of system states and, where applicable, physical parameters, pADAM supports forward prediction and inverse inference within a single architecture without retraining. Across benchmarks ranging from scalar diffusion to nonlinear Navier--Stokes equations, pADAM achieves accurate inference even under sparse observations. Combined with conformal prediction, it also provides reliable uncertainty quantification with coverage guarantees. In addition, pADAM performs probabilistic model selection from only two sparse snapshots, identifying governing laws through its learned generative representation. These results highlight the potential of generative multi-physics modeling for unified and uncertainty-aware scientific inference.
}

\maketitle


The mathematical description of physical phenomena through partial differential equations (PDEs) forms a cornerstone of modern science~\cite{evans2010pde,logan2014applied}, enabling the
characterization of systems ranging from large-scale weather and
climate dynamics~\cite{fisher2009data} to the complex turbulence of
fluid flows~\cite{foias2001navier,pletcher2013cfmht}. While classical numerical methods remain the bedrock of accuracy through systematic discretization~\cite{brenner2008fem, leveque2007fdm}, their extreme computational cost in high-dimensional parameter spaces—specifically in uncertainty quantification and inverse design—has motivated the
development of accelerated surrogate models~\cite{ghattas2021learning}.

Deep learning methods have emerged as a promising alternative to traditional PDE solvers, with architectures such as Fourier Neural Operators (FNO)~\cite{li2020fourier,kovachki2023neural, WEN2022104180, JIANG2024110392, LI2025117732}, Deep Operator Networks (DeepONet)~\cite{lu2021learning, HOWARD2023112462, ZHENG2025110329,yin2024scalable,jin2022mionet}, and Physics-Informed Neural Networks (PINNs)~\cite{RAISSI2019686, karniadakis2021physics} demonstrating the potential to bypass traditional solver constraints. Despite these advances, the field remains hindered by two critical limitations. First, most models produce deterministic outputs, limiting their utility in uncertainty quantification unless augmented with auxiliary techniques such as Bayesian training or ensemble methods~\cite{LIN2023111713, MOYA2025134418, LONE2026118552}. Second, and more fundamentally, current architectures operate within a ``one-model-one-equation'' paradigm. This rigidity prevents cross-physics knowledge transfer and necessitates exhaustive retraining for every new physical regime and task. Furthermore, while emerging foundation models for PDEs~\cite{ZHANG2024117229,LIU2024106707,sun2025towards,mccabe2024multiple,hao2024dpotautoregressivedenoisingoperator} have begun to address multi-operator training, they remain limited in
performing reliable inference under highly sparse observations
characteristic of real-world sensing. This limitation arises because these frameworks learn fixed forward mappings between prescribed input/output fields, rather than probabilistic models capable of being conditioned on arbitrary measurement operators at test time.

Diffusion models have recently emerged as a powerful generative
framework for scientific computing, demonstrating particular promise in solving PDEs by
learning the full probability distribution of solution states rather than deterministic
point estimates~\cite{huang2024diffusionpde, yao2025guided, wang2025fundiff, shan2025red, bastek2024physics}. Unlike deterministic operators, their iterative denoising formulation supports conditional sampling (inpainting), positioning them as inherently suited for inference under sparse observations. Despite this flexibility, existing diffusion-based PDE solvers remain constrained by the same specialization observed in earlier neural architectures: models are typically restricted to a single PDE class. Consequently, current approaches cannot generalize across disparate physics, perpetuating a fragmented landscape that limits their practical utility in settings involving heterogeneous physical regimes.

To address this fragmentation, we introduce pADAM (plug-and-play all-in-one diffusion architecture for multi-physics learning), a unified generative framework for learning across heterogeneous PDE families within a single model. pADAM learns a shared class-conditional probabilistic prior over system states and, when applicable, physical parameters, allowing forward prediction, parameter inference, and initial-condition reconstruction to be formulated within one posterior-sampling framework without task-specific retraining. In this way, a single architecture can operate across multiple physical regimes that are typically treated as separate learning problems. Through observation-guided conditional sampling, the model can incorporate sparse measurements at inference time, supporting accurate inference from only 10--30\% observations.

As a generative framework, pADAM also quantifies predictive uncertainty through sampling. To ensure the reliability of these uncertainty estimates, we integrate conformal prediction into the inference pipeline. This provides distribution-free finite-sample coverage guarantees for predictive intervals~\cite{vovk2005algorithmic, angelopoulos2021gentle,MOYA2025134418}, which is particularly important under sparse observations, where intervals can otherwise exhibit undercoverage. By calibrating predictive uncertainty, this framework supports more reliable scientific inference.

Beyond forward and inverse inference, pADAM also supports probabilistic PDE model selection from as few as two sparse temporal snapshots by leveraging its learned shared generative prior. This enables identification of governing physical laws and quantification of associated uncertainties from minimal measurements.

\begin{figure}[t]
    \centering
    \includegraphics[width=\linewidth]{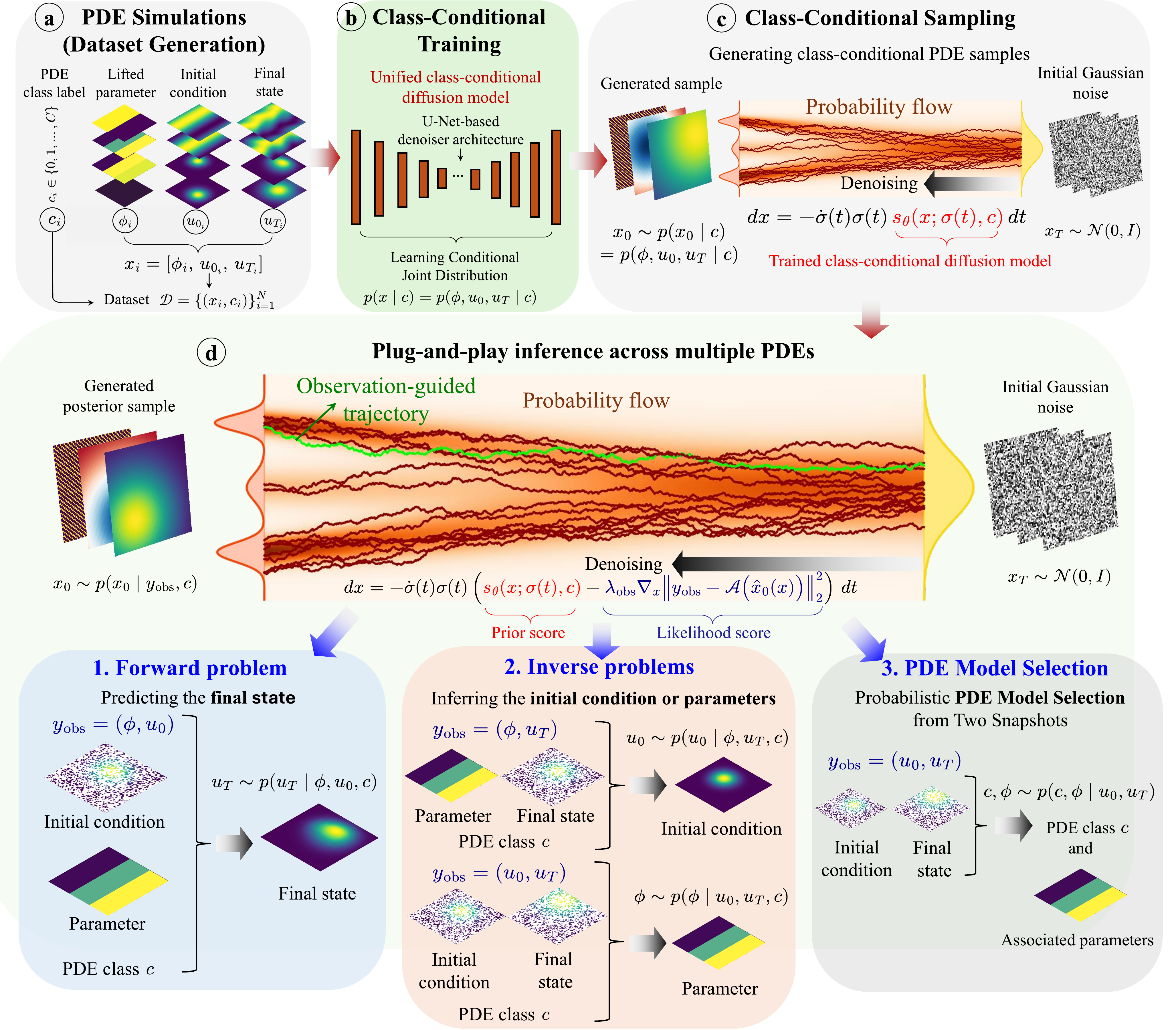} 
    \caption{\textbf{Schematic of the pADAM framework for unified multi-physics learning.} \textbf{a}--\textbf{c}, The pADAM framework learns across disparate physical laws, illustrated here for scalar-field PDEs, by projecting heterogeneous equation families into a shared generative prior. A class-conditional diffusion model learns the joint distribution of system states and physical parameters (\textbf{a, b}), enabling the generation of diverse physical regimes from Gaussian noise (\textbf{c}, orange trajectories). \textbf{d}, Task-agnostic inference via Bayesian conditioning. By incorporating full or sparse observations through plug-and-play guidance (green trajectory), the shared pADAM prior supports forward prediction, initial condition reconstruction, parameter inference, and probabilistic model selection within a single framework. This unified manifold allows pADAM to navigate a range of inference tasks without task-specific retraining.}
    \label{fig:schematic}
\end{figure}

Fig.~\ref{fig:schematic} provides a schematic overview of the pADAM architecture and its ability to transition across forward, inverse, and discovery tasks within a single probabilistic framework. This versatility positions pADAM as a promising framework for unified and uncertainty-aware scientific inference across heterogeneous physical systems.

\section*{Results}\label{sec2}

We evaluate pADAM as a unified foundation for multi-physics learning across a diverse library of heterogeneous PDEs. Our assessment consists of five investigations, each designed to evaluate a distinct capability of the framework. Across these investigations, we examine task-agnostic inference in forward and inverse problems under full and sparse observations, uncertainty quantification with conformal calibration, out-of-distribution extrapolation, and probabilistic model selection from minimal observations. Governing equations are detailed in the Methods, and the ranges of initial conditions and physical parameters across all evaluated PDE families are summarized in Extended Data Table~\ref{table:IC-Parameters}.

\subsection*{Unified multi-physics learning and trustworthy inference}
\label{subsec:foundation}

We first evaluate the capacity of pADAM to learn three distinct physical operators within a single architecture: purely dissipative (diffusion), purely convective (advection), and their coupled form (advection--diffusion). By unifying these heterogeneous PDE families---representing the transition from isotropic smoothing to directional transport---within a single set of weights, we assess whether a shared generative prior can capture disparate physical dynamics without equation-specific training.

\paragraph{Operator compression and parameter efficiency} To test this unification, we benchmark pADAM against PDE-specific diffusion-model baselines~\cite{huang2024diffusionpde} trained independently for each PDE family. In this configuration, physical parameters are held fixed to evaluate the model's ability to learn the bidirectional mapping between initial and final states. While the baselines require separate model instances for each regime, pADAM is trained to learn the conditional joint distribution $p(u_0, u_T \mid c)$, where $c$ denotes the PDE class. Crucially, pADAM uses approximately the same number of trainable parameters as a single specialized baseline, while supporting all three PDE families within a single model, for both forward prediction $u_T \sim p(u_T \mid u_0,c)$ and initial-state reconstruction $u_0 \sim p(u_0 \mid u_T,c)$ under full and sparse observations.

As reported in Table~\ref{table:exp1}, pADAM achieves high-fidelity predictions across all regimes, even under sparse observations. A notable result is the model's data efficiency; despite being trained on $33\%$ fewer samples per PDE family than the specialized baselines ($333$ vs. $500$ samples), pADAM achieves comparable accuracy overall across the evaluated tasks. This result is particularly important because pADAM integrates multiple dynamics within a single network, whereas standard approaches require independent model instances for each operator. By unifying these PDE families within a single model, pADAM demonstrates parameter efficiency, effectively amortizing the representational cost of the entire multi-physics library across a single set of weights. This operator compression suggests that parameter sharing induces a \emph{``shared physical vocabulary,''} enabling a single model to span multiple physical manifolds without increasing global computational overhead.

\newcommand{\best}[1]{\bfseries\num{#1}}

\sisetup{
  detect-weight = true,
  detect-family = true
}

\begin{table}[t]
\centering
\caption{\textbf{Comparative performance against PDE-specific baselines.} Relative $L_2$ error (\%) for forward prediction ($u_T$) and inverse reconstruction ($u_0$) across fixed-parameter PDE systems. We compare pADAM (trained on a diverse diffusion, advection, and advection--diffusion library, $N=1,000$) against PDE-specific diffusion baselines ($N=500$ per family) following DiffusionPDE~\cite{huang2024diffusionpde}. Results are averaged over 50 test instances under full (100\%) and sparse (30\%) spatial observations. The lowest errors in each row are shown in bold.}

\renewcommand{\arraystretch}{1.2}
\small
\setlength{\tabcolsep}{6pt}

\begin{tabular}{lllSS}
\toprule
\textbf{PDE system} & \textbf{Observation} & \textbf{Model} &
\multicolumn{1}{c}{\textbf{Forward ($u_T$)}} &
\multicolumn{1}{c}{\textbf{Inverse ($u_0$)}} \\
\midrule

\multirow{4}{*}{Diffusion}
 & \multirow{2}{*}{Full (100\%)}   
 & pADAM       & \num{0.82} & \best{0.45} \\
 &                   
 & Single-PDE   & \best{0.55} & \num{0.68} \\

 \cmidrule(lr){2-5}
 \addlinespace[0.1em]
 & \multirow{2}{*}{Sparse (30\%)}  
 & pADAM       & \best{0.88} & \best{0.86} \\
 &                   
 & Single-PDE   & \num{1.00} & \num{2.31} \\

\midrule

\multirow{4}{*}{Advection}
 & \multirow{2}{*}{Full (100\%)}   
 & pADAM       & \num{1.03} & \best{1.25} \\
 &                   
 & Single-PDE   & \best{1.02} & \num{1.26} \\

 \cmidrule(lr){2-5}
 \addlinespace[0.1em]
 & \multirow{2}{*}{Sparse (30\%)}  
 & pADAM       & \best{1.36} & \num{2.29} \\
 &                   
 & Single-PDE   & \num{1.52} & \best{1.55} \\

\midrule

\multirow{4}{*}{Advection--diffusion}
 & \multirow{2}{*}{Full (100\%)}   
 & pADAM       & \num{1.54} & \num{1.43} \\
 &                   
 & Single-PDE   & \best{1.28} & \best{1.29} \\

 \cmidrule(lr){2-5}
 \addlinespace[0.1em]
 & \multirow{2}{*}{Sparse (30\%)}  
 & pADAM       & \num{2.26} & \best{1.92} \\
 &                   
 & Single-PDE   & \best{1.64} & \num{2.05} \\

\bottomrule
\end{tabular}
\label{table:exp1}
\end{table}

\paragraph{Representation sharing across PDE operators}

To gain insight into how pADAM shares representations across heterogeneous operators, we examine its internal attention patterns. Analysis of a decoder block, shown in Extended Data Fig.~\ref{fig:exp1:attention_mechanism}, suggests that the model reuses and composes operator-specific features when transitioning from pure to mixed dynamics. The attention patterns for pure diffusion and pure advection are visibly distinct; in regions where attention weights are prominent for advection but absent for diffusion, the advection--diffusion maps exhibit intermediate values. This graded behavior is consistent with the mixed nature of the advection--diffusion equation, where transport and dissipation coexist. These patterns suggest that pADAM composes operator-specific features rather than maintaining isolated mechanisms for each PDE family.

To assess this behavior quantitatively, we compute pairwise cosine similarities between attention maps extracted from one encoder block and one decoder block across two denoising steps (Extended Data Fig.~\ref{fig:exp1:attention_similarity}). Across all settings, similarities involving the advection--diffusion case are consistently higher than those between the pure diffusion and pure advection extremes. This hierarchy mirrors the mathematical structure of the PDEs, where the mixed operator links the two pure regimes. Such alignment suggests that pADAM organizes its latent space in accordance with these relationships, reflecting the compositional structure of the underlying dynamics.

\paragraph{Zero-shot extrapolation to unseen physical laws}

A defining capability of a generalist prior is extrapolation beyond its training library. We evaluate pADAM—trained on diffusion, advection, and advection--diffusion—on the unseen advection--diffusion--reaction (ADR) equation, where the ADR system uses the same viscosity and velocity parameters as the advection--diffusion case. The reaction term ($k$) induces a structured operator shift absent during training, rendering the learned prior misspecified with respect to the true ADR dynamics. We quantify this shift using $\Delta_{\mathrm{op}}$, defined as the relative $L_2$ deviation of the terminal state induced by the reaction term (see Methods, Eq.~\eqref{eq:op_shift}). As shown in Extended Data Fig.~\ref{fig:adr_extrap_combined}a, the shift exceeds $20\%$ at $k=5.0$ and exceeds $50\%$ at $k=15.0$, confirming a genuinely out-of-distribution regime.

Despite this mismatch, pADAM remains robust when sparse observations of $(u_0, u_T)$ are used to jointly reconstruct the full-field states, as demonstrated in Extended Data Fig.~\ref{fig:adr_extrap_combined}b. The model leverages observation-guided sampling conditioned on the closest known operator class (the advection--diffusion class) to steer the misspecified prior toward trajectories consistent with the unseen ADR dynamics. While reconstruction error in $u_T$ increases with the reaction rate $k$, reflecting the accumulated influence of the unseen dynamics, errors for the initial state $u_0$ remain stable. As observation sparsity increases, errors rise in a controlled manner for both endpoints because less information is available to constrain the conditional distribution. These results suggest that pADAM can adapt to unseen dynamics through informative observations, supporting inference on previously unseen regimes without additional training.

\paragraph{Reliable uncertainty quantification via conformal calibration}
Unlike deterministic surrogates, the generative nature of pADAM provides a probabilistic basis for uncertainty quantification through posterior ensembles that capture distributions of physically consistent solutions. Although these ensembles reflect predictive uncertainty, empirical coverage (PICP) can fall well below the nominal 95\% target due to data limitations or the inherent ill-posedness that intensifies under observational sparsity (Extended Data Fig.~\ref{fig:exp1:PICP_vs_samples}). To address this, we integrate conformal prediction into pADAM to provide formal coverage guarantees and improve interval reliability.

As illustrated in Extended Data Fig.~\ref{fig:exp1:conformal_effect}, conformal calibration compensates for unresolved uncertainty by adaptively expanding prediction intervals in regions where observations provide limited information. Quantitative analysis in Extended Data Table~\ref{table:exp1_conformal} shows that, under 30\% spatial observations for the advection--diffusion system, mean empirical coverage increases from 58.33\% to 98.42\% for forward prediction and, most notably, from 36.31\% to 99.83\% for inverse reconstruction. The slight over-coverage relative to the 95\% target is consistent with finite-sample effects, as calibration set sizes are limited by sampling cost. Together, these results show that pADAM supports uncertainty-aware scientific inference with calibrated coverage even under severe data scarcity.

\subsection*{Navigating the continuous physics manifold}
\label{subsec:variable_phi}

To evaluate pADAM's capacity to model the continuous spectrum of physical dynamics, we extended the framework to systems with variable physical coefficients $\phi$. We trained a unified prior on three canonical PDE families, in which one physical coefficient was treated as a variable parameter for each system: diffusion (variable viscosity $\nu$), advection (variable velocity $a_x$ in the $x$-direction), and advection--diffusion (variable $\nu$). By learning the joint distribution $p(\phi, u_0, u_T \mid c)$, pADAM moves beyond discrete operator sets to represent a continuous physical manifold. This formulation enables task-agnostic inference, in which a single set of weights supports forward prediction $u_T \sim p(u_T \mid u_0, \phi,c)$, initial-state reconstruction $u_0 \sim p(u_0 \mid u_T, \phi,c)$, and parameter discovery $\phi \sim p(\phi \mid u_0, u_T,c)$ across disparate PDEs.

As summarized in Extended Data Table~\ref{ext_table:vp}, pADAM maintained high-fidelity performance across this full task spectrum; relative $L_2$ errors for both state and parameter estimation remained below 2.81\% under full observation. Notably, the model also remained stable under severe observational sparsity. While parameter inference is naturally more sensitive to sparsity-induced ill-posedness, state reconstruction errors remained below 4.11\% even with only 10\% spatial observations. These results suggest that pADAM can steer the generative prior toward physically consistent trajectories, supporting probabilistic state and parameter inference across continuous physical regimes. A qualitative illustration of parameter discovery on the continuous manifold for the advection system under sparse (10\%) observations is provided in Extended Data Fig.~\ref{fig:composite_results}a, further demonstrating pADAM's inference capability under limited data.

\subsection*{Scalability and generalization across the physical spectrum}
\label{subsec:scaling}

To evaluate the robustness of the pADAM framework, we investigated its scalability across two distinct dimensions: \emph{structural breadth} and \emph{parametric depth}. This dual-pronged assessment evaluates the model’s ability to maintain high-fidelity representations as both the diversity of governing laws and the dimensionality of their parameter spaces increase, providing a systematic evaluation of generalization across the physical spectrum.

\subsubsection*{Breadth: structural scaling to a 6-PDE library}

We next investigated the framework's capacity to scale to a broader training library by significantly expanding its structural breadth. To evaluate the representational efficiency of the learned manifold, we utilized a model with the same capacity and architecture as in the continuous physics manifold investigation and trained it on an expanded set of six PDE families, spanning both scalar and vector-valued systems. This structural scaling challenges the task-agnostic inference capabilities of pADAM by requiring it to learn a broader range of physical dynamics within a single set of weights.

For the scalar regimes---including diffusion, advection, advection--diffusion, and Allen--Cahn---we maintained the single-variable coefficient settings established in the continuous physics manifold investigation to evaluate whether the model could preserve its precision across the full task spectrum: forward prediction $u_T \sim p(u_T \mid u_0, \phi, c)$, initial-state reconstruction $u_0 \sim p(u_0 \mid u_T, \phi, c)$, and parameter discovery $\phi \sim p(\phi \mid u_0, u_T, c)$. For the vector-valued Burgers' and Navier--Stokes systems, the physical coefficients were held fixed to isolate the challenge of modeling coupled velocity fields via the joint distributions \( p(u_0, v_0, u_T \mid c) \) and \( p(u_0, v_0, v_T \mid c) \), which capture the relationship between the velocity components across initial and terminal states through shared conditioning. Here, pADAM models the coupled state transition through component-wise sampling; specifically, for forward prediction, we sample terminal velocities $u_T \sim p(u_T \mid u_0, v_0, c)$ and $v_T \sim p(v_T \mid u_0, v_0, c)$. For inverse state estimation, where joint reconstruction is inherently ill-posed, we leverage an auxiliary conditioning scheme---sampling $u_0 \sim p(u_0 \mid v_0, u_T, c)$ and $v_0 \sim p(v_0 \mid u_0, v_T, c)$---to better constrain the solution space and mitigate the sensitive dependence on initial conditions characteristic of nonlinear convective systems.

\begin{figure}[t!]
\centering

\begin{subfigure}{\linewidth}
    \centering
    \includegraphics[width=\linewidth]{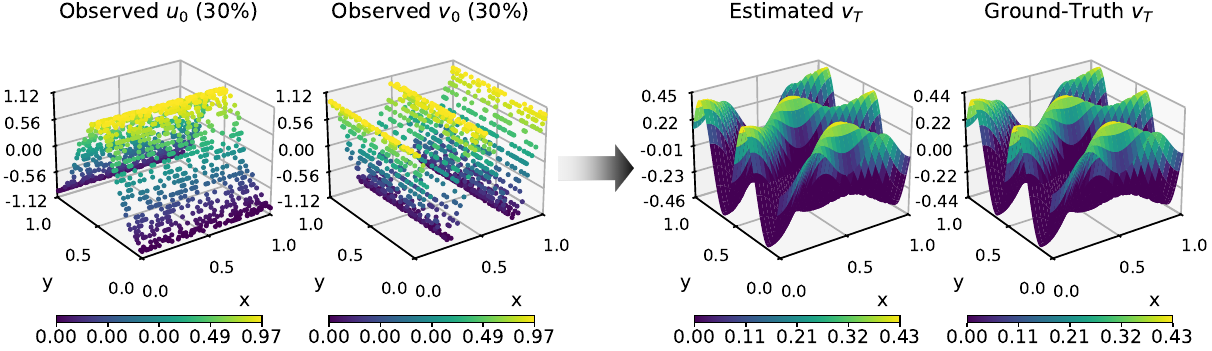}
    \subcaption{}
    \label{fig:navier_a}
\end{subfigure}

\vspace{6pt}

\begin{subfigure}{\linewidth}
    \centering
    \includegraphics[width=\linewidth]{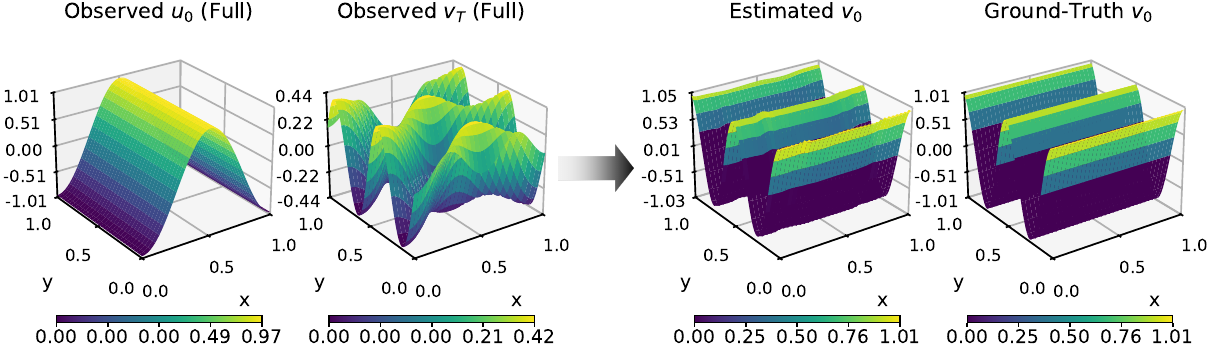}
    \subcaption{}
    \label{fig:navier_b}
\end{subfigure}

\vspace{6pt}

\begin{subfigure}{\linewidth}
    \centering
    \includegraphics[width=\linewidth]{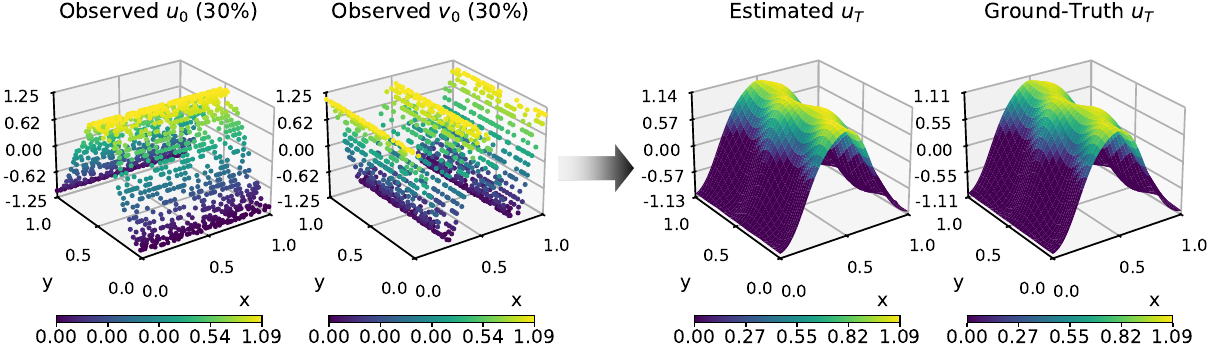}
    \subcaption{}
    \label{fig:navier_c}
\end{subfigure}

\caption{
\textbf{Forward and inverse inference for Navier--Stokes dynamics using pADAM.}
\textbf{a,} Forward prediction of the velocity component $v_T$ conditioned on 30\% spatial observations of the initial states $(u_0, v_0)$.
\textbf{b,} Inverse reconstruction of the initial velocity component $v_0$ conditioned on full observation of the terminal component $v_T$ and the initial component $u_0$.
\textbf{c,} Forward prediction of the velocity component $u_T$ conditioned on 30\% spatial observations of the initial states $(u_0, v_0)$.
}

\label{fig:Navier_Stokes}
\end{figure}

As reported in Table~\ref{table:exp3_structural}, pADAM maintained strong performance across the expanded library without architectural modification. While we observe a marginal increase in relative error compared to the smaller operator sets used in that earlier investigation, this limited degradation---despite the substantial increase in structural breadth---suggests that the model can maintain performance as the training library expands. Notably, the unified prior consistently captured disparate dynamic regimes and structural patterns, from the sharp interfaces of Allen--Cahn (retaining $<1\%$ parameter error) to the dissipative evolution of the Burgers' system. While the advection--diffusion system exhibited higher sensitivity in parameter discovery under sparsity ($14.36\%$), this reflects the compounded difficulty of resolving competing transport mechanisms within a shared latent space. In the Navier--Stokes regime, although the $v$-velocity component showed a characteristic error increase ($8.67\%$ at $30\%$ observations), this behavior is expected under sparse observations, where recovery of coupled nonlinear flow fields becomes more underdetermined. Nevertheless, the model remained stable and generated physically consistent samples. Qualitative examples of forward and inverse reconstructions for the Navier--Stokes and Burgers' equations are illustrated in Fig.~\ref{fig:Navier_Stokes} and Extended Data Fig.~\ref{fig:composite_results}b,c, respectively, highlighting the framework’s robust performance under both full and sparse observation regimes.


\begin{table}[h!]
\centering
\caption{
\textbf{Structural scaling across the multi-physics spectrum.} Relative $L_2$ errors (\%) for pADAM evaluated on a library of six distinct physical regimes. 
\textbf{a,} Performance on scalar-field PDEs (diffusion, advection, advection--diffusion, and Allen--Cahn) for forward prediction ($u_T$), initial-state reconstruction ($u_0$), and parameter discovery ($\phi$). 
\textbf{b,} Performance on vector-field PDEs (Burgers' and Navier--Stokes) where the model learns component-wise joint distributions to provide forward ($u_T, v_T$) and inverse ($u_0, v_0$) state estimation. 
All tasks are performed under both full (100\%) and sparse (30\%) spatial observations, with results reported as the mean over 50 test instances.
}
\label{table:exp3_structural}

\renewcommand{\arraystretch}{1.2}
\setlength{\tabcolsep}{2pt}
\small

\begin{tabular*}{\linewidth}{@{\extracolsep{\fill}}l l c c c c@{}}
\multicolumn{6}{c}{\textbf{a | Scalar-field PDEs}} \\
\midrule
\textbf{PDE system} & \textbf{Observation} & \textbf{Forward ($u_T$)} & \textbf{Inverse ($u_0$)} & \textbf{Inverse ($\phi$)} & \multicolumn{1}{c}{} \\
\midrule
\multirow{2}{*}{Diffusion}        & Full (100\%)   & 1.37 & 1.11 & 4.13 &  \\
                                  & Sparse (30\%)  & 2.03 & 2.40 & 5.60 &  \\
\midrule
\multirow{2}{*}{Advection}        & Full (100\%)   & 1.93 & 1.96 & 1.51 &  \\
                                  & Sparse (30\%)  & 1.93 & 1.97 & 2.24 &  \\
\midrule
\multirow{2}{*}{Advection--diff.} & Full (100\%)   & 1.96 & 2.28 & 9.26 &  \\
                                  & Sparse (30\%)  & 2.59 & 3.28 & 14.36 & \\
\midrule
\multirow{2}{*}{Allen--Cahn}      & Full (100\%)   & 1.21 & 2.42 & 0.48 &  \\
                                  & Sparse (30\%)  & 2.32 & 2.75 & 0.72 &  \\
\midrule

\addlinespace[6pt]
\multicolumn{6}{c}{\textbf{b | Vector-field PDEs}} \\
\midrule
\multicolumn{1}{c}{\textbf{PDE system}} &
\multicolumn{1}{c}{\textbf{Observation}} &
\shortstack{\textbf{Forward} \\ \textbf{($u_T$)}} &
\shortstack{\textbf{Forward} \\ \textbf{($v_T$)}} &
\shortstack{\textbf{Inverse} \\ \textbf{($u_0$)}} &
\shortstack{\textbf{Inverse} \\ \textbf{($v_0$)}} \\
\midrule
\multirow{2}{*}{Burgers'}       & Full (100\%)   & 1.36 & 1.16 & 1.29 & 0.90 \\
                                  & Sparse (30\%)  & 2.08 & 1.63 & 2.32 & 1.06 \\
\midrule
\multirow{2}{*}{Navier--Stokes}& Full (100\%)   & 1.45 & 4.96 & 1.25 & 3.85 \\
                                  & Sparse (30\%)  & 2.72 & 8.67 & 1.38 & 4.33 \\
\bottomrule
\end{tabular*}
\end{table}

\subsubsection*{Depth: parametric scaling and partial-parameter inference}\label{subsec:heterogeneous_results}

We further challenged the framework by moving from settings with a single variable parameter to regimes with multi-variable coefficient sets of different dimensionalities. This setting enables the evaluation of state and parameter inference in higher-dimensional parameter spaces. We trained pADAM on three canonical families—diffusion, advection, and advection--diffusion—in which all physical coefficients were treated as variable: $\phi = [\nu]$ for diffusion, $\phi = [a_x, a_y]$ for advection, and $\phi = [\nu, a_x, a_y]$ for advection--diffusion. To manage this heterogeneity, we employed a \emph{parameter lifting scheme} (see Methods) to project variable-length vectors and disparate physical parameters into a unified representation compatible with the conditional prior. We employed a model with the same capacity and architecture as in the continuous physics manifold and structural scaling investigations.

Under this increased parametric depth, pADAM performed state and multi-parameter inference across all families, as reported in Table~\ref{table:exp4_combined}a. Results indicate that pADAM maintained high accuracy for both forward prediction and inverse state reconstruction even as the parameter space expanded, while parameter inference exhibited higher error because of the increased ill-posedness associated with jointly inferring multiple coefficients. Qualitative examples of pADAM predictions across heterogeneous parameter dimensionalities are shown in Extended Data Fig.~\ref{fig:hetro}.

To demonstrate pADAM’s utility for partial-parameter inference, we leveraged the Bayesian guidance formulation to incorporate \emph{a priori} physical knowledge at inference time. For the advection system, we compared the joint inference of the velocity vector $\phi=[a_x, a_y]$ with conditional settings in which one component was treated as an observed constraint. As summarized in Table~\ref{table:exp4_combined}b, incorporating this \emph{a priori} knowledge---for example, $a_x \mid a_y$---substantially improved the estimation of the remaining component across both full and sparse observational regimes. By explicitly restricting the parameter manifold to configurations consistent with known physical constraints, pADAM mitigates ill-posedness by reducing the effective dimensionality of the inverse problem. This capability enables the integration of physical priors into real-time inference without model retraining.

\begin{table}[t!]
\caption{
\textbf{Parametric scaling and partial-parameter inference performance.} 
\textbf{a,} Task-agnostic inference across a heterogeneous PDE library with different parametric dimensionality. Relative $L_2$ errors (\%) for state estimation ($u_0$ and $u_T$) and parameter recovery ($\phi$) across three physical regimes with varying parametric depth: Diffusion ($\phi=[\nu]$), Advection ($\phi=[a_x, a_y]$), and Advection--diffusion ($\phi=[\nu, a_x, a_y]$). 
\textbf{b,} Impact of partial parameter constraints on parameter inference in the advection system. Comparison of joint inference ($a_x, a_y$) against conditional settings where one velocity component ($a_x$ or $a_y$) is provided at inference time. All tasks are performed under both full (100\%) and sparse (30\%) spatial observations, with results representing the mean over 50 test instances.
}
\label{table:exp4_combined}
\centering

\renewcommand{\arraystretch}{1.2}
\setlength{\tabcolsep}{2.3pt} 
\small

\begin{tabular*}{\linewidth}{@{\extracolsep{\fill}}l l c c c@{}}
\multicolumn{5}{@{}p{\linewidth}@{}}{\centering\textbf{a | Task-agnostic inference across a heterogeneous PDE library with different parametric dimensionality}} \\

\midrule
\textbf{PDE system} & \textbf{Observation} & \textbf{Forward ($u_T$)} & \textbf{Inverse ($u_0$)} & \textbf{avg. Inverse ($\phi$)} \\
\midrule
\multirow{2}{*}{Diffusion}         &Full (100\%) & 0.98 & 1.18 & 2.89 \\
                                   &Sparse (30\%)& 1.82 & 2.07 & 4.61 \\
\midrule
\multirow{2}{*}{Advection}         &Full (100\%) & 0.85 & 1.03 & 6.57 \\
                                   &Sparse (30\%)& 1.64 & 1.32 & 6.61 \\
\midrule
\multirow{2}{*}{Advection--diff.}  &Full (100\%) & 1.49 & 1.72 & 5.99 \\
                                   &Sparse (30\%)& 2.31 & 2.71 & 5.95 \\
\midrule
\addlinespace[12pt]
\multicolumn{5}{@{}p{\linewidth}@{}}{\centering\textbf{b | Partial-parameter inference under conditional physical constraints in the advection system}} \\
\midrule
\textbf{Observation} & \textbf{Inference setting} & \textbf{Inverse  ($a_x$)} & \textbf{Inverse ($a_y$)} & \\
\midrule
\multirow{3}{*}{Full (100\%)}
 & $a_x,a_y \sim p(a_x, a_y \mid u_0, u_T)$    & 6.31 & 6.83 &  \\
 & $a_x \sim p(a_x \mid a_y, u_0, u_T)$        & 2.00 & --   &  \\
 & $a_y \sim p(a_y \mid a_x, u_0, u_T)$        & --   & 2.30 &  \\
\midrule
\multirow{3}{*}{Sparse (30\%)}
 & $a_x,a_y \sim p(a_x, a_y \mid u_0, u_T)$   & 6.32 & 6.91 &  \\
 & $a_x \sim p(a_x \mid a_y, u_0, u_T)$       & 2.44 & --   &  \\
 & $a_y \sim p(a_y \mid a_x, u_0, u_T)$       & --   & 2.92 &  \\
\bottomrule
\end{tabular*}
\end{table}

\subsection*{Identifying governing laws through probabilistic model selection}
\label{subsec:discovery}

A central challenge in scientific machine intelligence is the autonomous identification of governing laws from sparse, temporal observations. We next evaluate pADAM's capacity for probabilistic model selection, challenging the framework to infer the underlying physical law and its associated coefficients from as few as two sparse snapshots ($c, \phi \sim p(c, \phi \mid u_0, u_T)$). pADAM performs identification by leveraging its learned class-conditional prior to explore a candidate library of PDE classes. For this evaluation, we utilize the generalist prior trained on the heterogeneous parameter datasets used in the parametric scaling investigation.

\begin{figure}[t]
    \centering
    
    \begin{minipage}[c]{0.44\textwidth} 
        \centering
        \includegraphics[width=\linewidth]{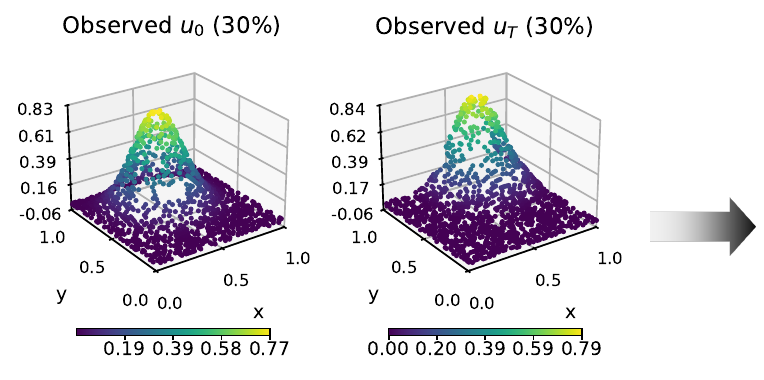}
        \vspace{2pt}
        \subcaption{\centering Advection regime}
    \end{minipage}
    \hfill
    \begin{minipage}[c]{0.54\textwidth} 
        \footnotesize
        \begin{equation*}
        \begin{aligned}
        \text{\textbf{True PDE:}} \,\, & \partial_t u + [2.56, 2.97] \cdot \nabla u = 0 \\
        \text{\textbf{Sampled PDE:}} \,\, & \partial_t u + [2.60, 2.87] \cdot \nabla u = 0 \\
        \text{\textbf{95\% Interval:}} \,\, & \partial_t u + [2.70 \pm 0.10, 2.79 \pm 0.09] \cdot \nabla u = 0 \\
        \end{aligned}
        \end{equation*}
    \end{minipage}

    \vspace{1.5em}

    \begin{minipage}[c]{0.44\textwidth}
        \centering
        \includegraphics[width=\linewidth]{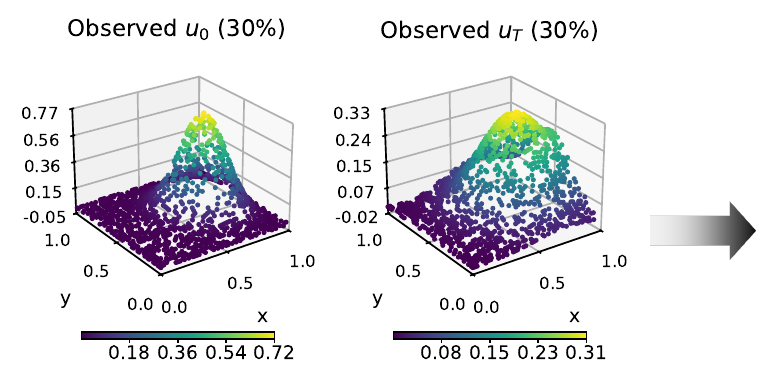}
        \vspace{2pt}
        \subcaption{\centering Diffusion regime}
    \end{minipage}
    \hfill
    \begin{minipage}[c]{0.54\textwidth}
        \footnotesize
        \begin{equation*}
        \begin{aligned}
        \text{\textbf{True PDE:}} \,\, & \partial_t u = 0.29 \Delta u \\
        \text{\textbf{Sampled PDE:}} \,\, & \partial_t u = 0.30 \Delta u \\
        \text{\textbf{95\% Interval:}} \,\, & \partial_t u = 0.30(\pm 0.0010) \Delta u \\
        \end{aligned}
        \end{equation*}
    \end{minipage}

    \vspace{1.5em}

    \begin{minipage}[c]{0.44\textwidth}
        \centering
        \includegraphics[width=\linewidth]{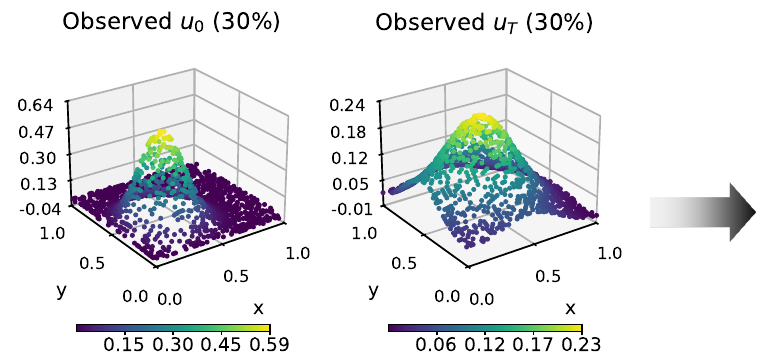}
        \vspace{2pt}
        \subcaption{\centering Advection--diffusion regime}
    \end{minipage}
    \hfill
    \begin{minipage}[c]{0.54\textwidth}
        \footnotesize
        \begin{equation*}
        \begin{aligned}
        \text{\textbf{True PDE:}} \,\, & \partial_t u + [2.40, 2.30] \cdot \nabla u = 0.33 \Delta u \\
        \text{\textbf{Sampled PDE:}} \,\, & \partial_t u + [2.37, 2.19] \cdot \nabla u = 0.34 \Delta u \\
        \text{\textbf{95\% Interval:}} \,\, & \partial_t u + [2.29 \pm 0.06, 2.35 \pm 0.18] \cdot \nabla u \\
                                     & \qquad = 0.34(\pm 0.016) \Delta u \\
        \end{aligned}
        \end{equation*}
    \end{minipage}

    \caption{\textbf{Probabilistic model selection across three PDE classes.} Identification of governing laws from two snapshots $(u_0, u_T)$ with only 30\% of the spatial field observed. \textbf{a--c,} Left panels show sparse field snapshots; right panels compare the ground-truth PDEs with representative sampled PDEs and the associated 95\% predictive intervals derived from the ensemble.}
    
    \label{fig:exp5:discovery_combined}
\end{figure}

As illustrated in Fig.~\ref{fig:exp5:discovery_combined}, the framework reliably distinguishes between competing physical hypotheses—such as advective transport versus pure diffusion—even under significant observational sparsity. This selection follows an \emph{infer-and-validate} logic (see Methods): for each candidate class, pADAM leverages its shared generative prior to infer the corresponding coefficient posterior and assess marginal consistency by cross-referencing parameter inference with generative state reconstructions. This process allows the framework to compare candidate PDE classes and identify the governing law that best explains the observed state transition.

Despite severe observational scarcity (30\% observations), pADAM consistently selects the ground-truth operator across the evaluated scenarios while maintaining parameter estimates that closely align with the true coefficients. Because the framework is fully generative, repeated sampling naturally induces a distribution over candidate laws and associated parameters; the resulting ensemble-based predictive intervals provide a measure of the epistemic uncertainty inherent in physical model selection. These results suggest that pADAM can serve as a probabilistic framework for characterizing systems whose governing dynamics are ambiguous or only partially observed from limited data.

\section*{Discussion}
\label{sec:discussion}

The results show that a single class-conditional generative prior can support forward prediction, inverse inference, and probabilistic model selection across multiple PDE families, including both scalar and vector-valued systems. Rather than relying on separate task-specific solvers, pADAM learns a shared probabilistic representation that can be conditioned on different forms of partial information. This suggests that heterogeneous physical systems can be addressed within a unified generative framework while maintaining accurate inference under sparse observations.

A central result is that pADAM can identify governing laws from only two sparse snapshots by comparing how well candidate PDE classes explain the observed transition under the learned prior. Unlike classical discovery methods that rely on dense trajectories, this approach formulates PDE identification as probabilistic inference over a candidate library of physical models. Because the comparison is generative rather than deterministic, it also yields uncertainty over candidate laws and associated parameters. 

Crucially, conformal calibration strengthens pADAM as a framework for scientific inference by providing distribution-free finite-sample coverage guarantees for predictive intervals. This is particularly important in sparse and ill-posed settings, where raw predictive intervals may exhibit substantial undercoverage. In addition, the zero-shot extrapolation results suggest that the learned prior is not limited to a fixed training library and can remain effective under operator shift when guided by observations.

The pADAM framework is designed to remain compatible with a broad class of generative backbones, making it adaptable to emerging paradigms such as flow matching~\cite{lipman2022flow} and recent mean-flow approaches~\cite{geng2025mean}, which may improve sampling efficiency and reduce computational cost. Future extensions could further enhance performance, particularly in extreme low-data regimes, by incorporating physics-informed priors directly into the generative process~\cite{bastek2024physics}. In addition, extending the framework to utilize function-space diffusion~\cite{yao2025guided} may enable discretization-invariant inference in infinite-dimensional settings and support inference across varying grid resolutions.

Beyond these technical extensions, an important next direction is to move beyond fixed candidate libraries toward more flexible forms of equation discovery for systems with partially unknown physics. Such developments could improve the ability of generative frameworks to relate high-dimensional observations to interpretable mathematical structure.

Taken together, these results suggest that shared generative physical priors can support prediction, inference, and model discovery within a single probabilistic framework. More broadly, this work can be viewed as a pilot step toward foundation models for science, in which large generative models are trained across diverse physical domains and adapted to downstream scientific and engineering tasks. In that sense, pADAM points toward a class of generalist scientific machine learning models that remain effective even when observations are sparse, irregular, or incomplete.

\section*{Methods}

\subsection*{Problem formulation and objective}

We consider a library of $C$ distinct PDE families. Each class $c \in \{1,\dots,C\}$ is associated with a governing operator $\mathcal{F}^{(c)}$ representing a specific physical dynamical regime:
\begin{equation}
\mathcal{F}^{(c)}\!\big(u(\mathbf{s},t);\phi\big) = 0, \quad (\mathbf{s},t)\in\Omega\times(0,T], \quad u(\mathbf{s},0)=u_0(\mathbf{s}),
\end{equation}
subject to appropriate boundary conditions on $\partial\Omega$. Here, $u(\mathbf{s},t)$ denotes the solution field in spatial coordinates $\mathbf{s}\in\Omega$, and $\phi\in\mathbb{R}^{d_c}$ represents the vector of physical coefficients parameterizing the dynamics within class $c$. This formulation accommodates both scalar-valued and vector-valued states (for example, coupled velocity components $\mathbf{u}=(u,v)$ in Navier--Stokes). In this work, all experiments are conducted on two-dimensional spatial domains; thus, $\mathbf{s}\in\Omega\subset\mathbb{R}^2$.

The objective is to learn a unified, class-conditional generative prior $p(x \mid c)$, where $x$ is a multi-channel variable encapsulating the joint distribution of temporal system states and, where applicable, governing physical parameters within a shared probabilistic representation. By parameterizing this prior via a diffusion model~\cite{song2021scorebased, ho2020ddpm}, we treat forward prediction, inverse state estimation, and parameter recovery as posterior sampling tasks of the form $p(x \mid y_{\text{obs}}, c)$, where $y_{\text{obs}}$ represents an arbitrary set of observations spanning from full-field data to \emph{sparse} measurements. This formulation enables task-agnostic inference, in which diverse downstream physical tasks are performed by conditioning a shared manifold of physical dynamics on available data without requiring task-specific architectures or retraining.

\subsection*{The pADAM framework: a unified foundation for multi-physics learning}

The pADAM framework brings heterogeneous PDE families into a shared generative formulation (Fig.~\ref{fig:schematic}). As illustrated in Fig.~\ref{fig:schematic}a--c, a class-conditional diffusion model learns the joint distribution of the unified representation, comprising system states and, where applicable, physical parameters, and thereby supports generation across multiple physical regimes from Gaussian noise (Fig.~\ref{fig:schematic}c, orange trajectories). As shown in Fig.~\ref{fig:schematic}d, task-agnostic inference is performed through Bayesian conditioning: by incorporating full or sparse observations through plug-and-play guidance (green observation-guided trajectory), the shared pADAM prior supports forward prediction, inverse state and parameter inference, uncertainty quantification, and probabilistic model selection within a single framework. We detail each component of pADAM in the following sections.

\subsubsection*{Unified joint representation and parameter lifting}\label{joint_representation}

To enable a single architecture to process heterogeneous physics, we transform each system into a unified 3-channel generative variable $x \in \mathbb{R}^{3 \times N_x \times N_y}$. This representation ensures architectural invariance across different physical regimes:
\begin{equation}
x := \begin{cases} [\,\Phi,\,u_0,\,u_T\,] & \text{for scalar-field PDEs} \\ [\,u_0,\,v_0,\,u_T \text{ or } v_T\,] & \text{for vector-field PDEs} \end{cases}
\end{equation}
where $\Phi$ is a spatially lifted representation of the coefficients $\phi$, and $u, v$ represent coupled velocity-field components. By maintaining a fixed channel dimension, a single class-conditional diffusion model can be deployed across diverse dynamical systems without structural modification. While this configuration is optimized for the forward and inverse tasks presented in this work, the framework is inherently modular, allowing for alternative channel assignments depending on the requirements of the physical regime.

\noindent\textbf{Generative inclusion of physical parameters.} Treating $\phi$ as a component of the generative variable $x$, rather than a static conditioning input, is a key modeling choice for probabilistic inverse inference in parametric systems. This design allows the model to learn the intrinsic joint prior $p(\phi, u_0, u_T \mid c)$, enabling the recovery of unknown physical parameters through posterior sampling $\phi \sim p(\phi \mid u_0, u_T, c)$ at inference time.

\noindent\textbf{Mathematical lifting scheme.} To bridge the dimensionality gap between coefficients $d_c$ and spatial states $N_x \times N_y$, we define a lifting operator $\mathcal{L}$ that broadcasts $\phi$ into a spatial field $\Phi$. This process ensures that physical parameters are represented as spatially compatible features:
\begin{itemize}
    \item \textbf{Scalar case ($d_c=1$):} $\Phi_{i,j} = \phi$ for all $(i,j) \in \Omega$.
    \item \textbf{Vector case ($d_c > 1$):} $\Phi_{i,j} := \sum_{k=1}^{d_c} \phi_k \mathds{1}_{\Omega_k}(i,j)$, where $\{\Omega_k\}_{k=1}^{d_c}$ is a disjoint partition of the spatial grid.
\end{itemize}
This spatial lifting provides a consistent inductive bias, allowing the model's hierarchical convolutional architecture to capture the functional dependencies between the governing physical parameters ($\Phi$) and the system states ($u_0, u_T$).

\subsubsection*{Learning unified generative priors across heterogeneous physical regimes}

To enable a unified modeling framework for multi-physics systems, we learn a single generative prior $p(x \mid c)$ using a class-conditional diffusion model~\cite{JMLR:v23:21-0635} that captures the joint distribution of system states and, where applicable, governing physical parameters across the library of PDE families $\mathcal{C}$. This approach leverages shared latent structure across heterogeneous operators, establishing a unified generative representation that captures dynamical structure across disparate PDE families.

The representation is adapted to the structural dependencies of each physical family:
\begin{itemize}
    \item \textbf{Parametric scalar-field PDEs:} We learn the joint prior $p(\Phi, u_0, u_T \mid c)$. By including the lifted parameter field $\Phi$ in the generative variable $x$, the model captures the relationship between physical coefficients and system states.

    \item \textbf{Vector-valued PDEs:} For systems with fixed physical parameters, we learn component-wise joint priors, $p(u_0, v_0, u_T \mid c)$ and $p(u_0, v_0, v_T \mid c)$. This strategy uses the available channel capacity to resolve coupled velocity components, enabling the model to capture dependencies between initial and terminal states.
\end{itemize}

We adopt the EDM framework~\cite{karras2022elucidating} to parameterize this prior. In this implementation, the model is trained using a conditional denoising score-matching objective:
\begin{equation}
\mathcal{L}_{\text{train}} = \mathbb{E}_{x_0, c, \sigma, n} \left[ \lambda(\sigma) \| D_\theta(x_0 + n; \sigma, c) - x_0 \|_2^2 \right],
\label{eq:loss_edm}
\end{equation}

where $n \sim \mathcal{N}(0,\sigma(t)^2 \mathbf{I})$ denotes additive Gaussian noise at diffusion level $\sigma(t)$, and $\lambda(\sigma)$ is a preconditioning weight. The diffusion process defines a time-dependent generative variable $x=x(t)$ for $t \in [0,T]$, where $x_0=x(0)$ denotes a clean sample from the class-conditional data distribution and $x(t)$ becomes progressively noisier as $t$ increases under the noise schedule $\sigma(t)$. At the terminal time $t=T$, the diffusion distribution approaches a Gaussian prior, so that sampling begins from a Gaussian noise distribution and is then transported back toward the data manifold. Here, $D_\theta$ denotes the denoiser that maps noisy inputs toward the clean manifold. Through the relation 
$s_\theta(x; \sigma(t), c) = (D_\theta(x; \sigma(t), c) - x)/{\sigma(t)}^2$, 
the network learns to approximate the conditional score function, yielding an estimate of the log-density gradient $\nabla_x \log p_t(x\mid c)$, where $p_t(x \mid c)$ denotes the class-conditional marginal distribution of the diffusing sample at time $t$. Once this score estimator is obtained, samples are evolved through a deterministic mapping governed by the Probability Flow Ordinary Differential Equation (ODE)~\cite{song2021scorebased,karras2022elucidating}:

\begin{equation}
\frac{d x}{dt}
=
-\dot{\sigma}(t)\,\sigma(t)\,
s_\theta\!\left(x;\sigma(t),c\right),
\label{eq:prob_flow_ode}
\end{equation}

By evolving this ODE at inference time, pADAM transforms stochastic noise into physically consistent realizations, steering sample trajectories along the learned class-conditional manifold so that generated samples remain consistent with the characteristic dynamics of the specified physical regime.

\subsubsection*{Task-agnostic inference via plug-and-play observational guidance}

A key feature of pADAM is its ability to perform task-agnostic inference in a \emph{plug-and-play} manner, where forward prediction, inverse state estimation, and coefficient recovery are unified as conditional sampling problems without requiring model retraining. Given full or sparse observations $y_{\text{obs}}$, we sample from the posterior distribution $p(x \mid y_{\text{obs}}, c)$ using the learned class-conditional generative prior. Following Bayes' rule, the posterior score is decomposed as~\cite{song2021scorebased}:

\begin{equation}
\nabla_{x} \log p_{t}(x \mid y_{\text{obs}}, c)
\approx
\underbrace{s_\theta(x; \sigma(t), c)}_{\text{Prior Score}}
+
\underbrace{\nabla_{x} \log p_t(y_{\text{obs}} \mid x, c)}_{\text{Likelihood Score}},
\label{eq:score_decomp}
\end{equation}
where the prior score is provided by the pre-trained class-conditional model, and the likelihood score enforces consistency with available measurements.

Under additive Gaussian measurement noise, and following the formulation for score-based guidance~\cite{chung2022diffusion, huang2024diffusionpde}, the likelihood score is approximated at inference time as:

\begin{equation}
\nabla_{x} \log p_t(y_{\text{obs}} \mid x, c)
\approx
-\lambda_{\mathrm{obs}}
\nabla_{x}
\left\|
y_{\text{obs}}
-
\mathcal{A}\!\left(\hat{x}_0(x)\right)
\right\|_2^2,
\label{eq:likelihood_approx}
\end{equation}
where $\mathcal{A}$ denotes a task-specific measurement operator, and $\hat{x}_0$ is the denoiser’s estimate of the clean system representation. Substituting this approximation into the Probability Flow ODE \eqref{eq:prob_flow_ode} yields the guided probability-flow dynamics:

\begin{equation}
\frac{d x}{dt}
=
-\dot{\sigma}(t)\,\sigma(t)
\left(
s_\theta(x;\sigma(t),c)
-
\lambda_{\mathrm{obs}}
\nabla_{x}
\left\|
y_{\text{obs}}
-
\mathcal{A}\!\left(\hat{x}_0(x)\right)
\right\|_2^2
\right),
\label{eq:guided_ode}
\end{equation}
The guidance scale $\lambda_{\mathrm{obs}}$ acts as a weighting factor that balances the learned prior against observational constraints. Under this formulation, the guided ODE maps an initial Gaussian noise sample to realizations that remain consistent with both the specified physical class and the available measurements. The prior score steers the trajectory toward high-probability regions of the class-conditional distribution, while the observation-based correction enforces consistency with $y_{\text{obs}}$.

\paragraph{Forward and inverse operator synthesis}

The flexibility of the observation operator $\mathcal{A}$ enables pADAM to support inference across a diverse set of physical operators within a single architecture. This versatility is demonstrated across three fundamental classes of physics problems:

\begin{itemize}
    \item \textbf{Forward prediction:} For parametric scalar PDEs, given the coefficients $\Phi$ and the initial state $u_0$, pADAM samples the terminal state $u_T \sim p(u_T \mid u_0, \Phi, c)$. For vector-valued systems (e.g., Navier--Stokes), the model performs component-wise prediction of terminal velocities $u_T \sim p(u_T \mid u_0, v_0, c)$ and $v_T \sim p(v_T \mid u_0, v_0, c)$.
    
    \item \textbf{Inverse state estimation:} For scalar fields, pADAM samples $u_0 \sim p(u_0 \mid u_T, \Phi, c)$. For multi-component systems where joint reconstruction is highly ill-posed, we sample $u_0 \sim p(u_0 \mid v_0, u_T, c)$ and $v_0 \sim p(v_0 \mid u_0, v_T, c)$; this auxiliary conditioning constrains the solution space and mitigates the ill-posedness of the inverse recovery.

    \item \textbf{Inverse parameter identification:} For scalar parametric systems, given paired observations of the initial and terminal states $(u_0, u_T)$, pADAM recovers the governing coefficients $\Phi \sim p(\Phi \mid u_0, u_T, c)$.
    
\end{itemize}

Crucially, $\mathcal{A}$ can represent both full-field and sparse observations, allowing pADAM to operate across different data-density regimes. By unifying these tasks across heterogeneous physical domains, pADAM functions as a \emph{probabilistic framework for multi-operator, multi-physics inference}. This shared formulation enables forward and inverse problems to be addressed within a single architecture without task-specific retraining.

\subsubsection*{Reliable uncertainty quantification via conformal calibration}\label{subsec:uq}

As a generative framework, pADAM provides uncertainty quantification (UQ) through posterior sampling. By drawing multiple conditional samples from the observation-guided reverse process, we obtain an empirical distribution consistent with both the physical constraints of the PDE class and the available measurements. The uncertainty captured here is primarily \emph{epistemic}, reflecting ambiguity arising from factors including sparse observations, inverse ill-posedness, and limited data. This is particularly important in scientific settings where reliable uncertainty estimates are needed to support inference.

\paragraph{Ensemble-based uncertainty estimation}
For a given conditioning specification—encompassing the PDE class, partial state observations, and known coefficients—we generate an ensemble of $M$ independent samples $\{z_j\}_{j=1}^M$ from the guided reverse diffusion process~\eqref{eq:guided_ode}, where $z \in \{\Phi, u_0, u_T, v_0, v_T\}$. We characterize this predictive distribution through the ensemble mean $\mu_M$ and standard deviation $\sigma_M$:
\[
\mu_M = \frac{1}{M} \sum_{j=1}^M z_j, \qquad
\sigma_M = \sqrt{\frac{1}{M-1} \sum_{j=1}^M \bigl(z_j - \mu_M\bigr)^2}.
\]
While a standard Gaussian-based $95\%$ interval ($\mu_M \pm 1.96\sigma_M$) is a common baseline, conditional distributions in chaotic or under-determined PDE regimes frequently deviate from Gaussianity, exhibiting significant skewness, heavy tails, or multimodality. Consequently, these nominal intervals provide no formal guarantees and can suffer from substantial miscalibration. This issue is especially relevant in ill-posed inverse inference and prediction under sparse observations, where multiple physical solutions may be consistent with the available data.

\paragraph{Conformal calibration with distribution-free coverage}
To provide the reliability required for scientific inference, we integrate \emph{conformal prediction}~\cite{vovk2005algorithmic, angelopoulos2021gentle}. This procedure post-processes pADAM's ensemble estimates to produce prediction intervals with finite-sample, distribution-free coverage guarantees. Because the scale of uncertainty varies across heterogeneous physical regimes, calibration is performed independently for each class--task pair $(c, \tau)$. Under the exchangeability assumption for the calibration and test samples, we compute nonconformity scores $s_i^{(c,\tau)}$ on a calibration dataset $\mathcal{D}_{\mathrm{cal}}^{(c,\tau)}$ of size $n_{c,\tau}$ to quantify the normalized discrepancy between the ground truth and the predictive distribution~\cite{romano2019conformalized, MOYA2025134418}:
\begin{equation}
s_i^{(c,\tau)} = \frac{\lvert z_i - \mu_M(y_{\mathrm{obs},i}) \rvert}{\sigma_M(y_{\mathrm{obs},i})}.
\end{equation}
Given a target miscoverage rate $\alpha \in (0,1)$, the calibration threshold $\hat{q}_\alpha^{(c,\tau)}$ is defined as the $\lceil (n_{c,\tau}+1)(1-\alpha) \rceil / n_{c,\tau}$-th empirical quantile of the scores in $\mathcal{D}_{\mathrm{cal}}^{(c,\tau)}$. We then construct the final conformal interval:
\begin{equation}
\mathcal{C}^{(c,\tau)}_\alpha(y_{\mathrm{obs,test}}) = \Bigl[ \mu_M(y_{\mathrm{obs,test}}) \pm \hat{q}^{(c,\tau)}_\alpha \sigma_M(y_{\mathrm{obs,test}}) \Bigr].
\end{equation}
Under exchangeability, this construction guarantees marginal coverage of at least $1-\alpha$, providing a principled measure of predictive reliability.

\subsubsection*{Probabilistic PDE model selection from two snapshots} \label{sec:model_selection_two_snapshots}

A core strength of pADAM is its ability to accommodate heterogeneous parameter representations across multiple PDE families through parameter lifting. This architectural design maintains structural consistency even when the dimensionality of the physical parameters varies across PDE classes. For example, a PDE library may include diffusion equations defined by scalar diffusivity $\nu$, advection equations governed by velocity components $(a_x, a_y)$, and advection--diffusion systems parameterized by $(\nu, a_x, a_y)$. We leverage this capability to perform \emph{probabilistic model selection from only two snapshots} for scalar-field PDEs, casting the identification of governing laws from sparse temporal observations as a probabilistic inference task.

Given only a pair of state snapshots, consisting of an initial state $u_0$ and a terminal state $u_T$, which may be available through sparse measurements, the goal is to identify the governing PDE class from a candidate library $\mathcal{C} = {c_1, \dots, c_K}$ and infer the associated physical parameters. This setting is highly ill-posed and represents a form of probabilistic PDE discovery. In contrast to classical identification approaches requiring dense spatiotemporal trajectories, pADAM leverages its learned multi-operator prior to evaluate the consistency of each candidate law with the observed transition. By sampling class-conditional parameter posteriors, the framework effectively reconstructs the joint posterior $p(c, \phi \mid u_0, u_T)$, naturally quantifying the epistemic uncertainty inherent in identifying physics from limited temporal snapshots.

The selection procedure follows a \emph{infer-and-validate} logic:

\begin{enumerate}
    \item \textbf{Conditional parameter inference:} For each candidate class $c \in \mathcal{C}$, we sample from the class-conditional parameter posterior $\hat{\phi}^{(c)} \sim p(\phi \mid u_0, u_T, c)$. This identifies parameter configurations that are maximally consistent with the observed state transition under class $c$.
    
    \item \textbf{Generative validation:} We then use forward prediction to sample a synthetic terminal state $\hat{u}_T^{(c)} \sim p(u_T \mid \hat{\phi}^{(c)}, u_0, c)$. Each candidate is evaluated based on its generative reconstruction discrepancy:
    \begin{equation}
        \mathcal{E}(c) = \| \hat{u}_T^{(c)} - u_T \|_2.
    \end{equation}
\end{enumerate}

The identified operator $c^\star = \arg\min_{c \in \mathcal{C}} \mathcal{E}(c)$, together with the corresponding inferred coefficients $\phi^\star = \hat{\phi}^{(c^\star)}$, defines the explicit governing model that best explains the observed dynamics. Because pADAM is fully generative, repeated ensemble sampling allows us to quantify the uncertainty associated with model selection. This enables a principled approach to \emph{uncertainty-aware model selection}, where the relative support for competing physical laws is reflected in the distribution of reconstruction discrepancies.
\subsection*{Experimental design}

\subsubsection*{Benchmark PDE Problems for Experiments}
\label{sec:pde-datasets}

To assess the generality of the pADAM framework, we consider seven representative PDE families spanning dissipative, advective, mixed, and nonlinear dynamics. This multi-physics library provides a stringent testbed for unified operator learning across heterogeneous physical regimes.

Datasets for the diffusion, advection, advection--diffusion, advection--diffusion--reaction, Burgers', and Allen--Cahn equations are generated using finite-difference discretizations, while Navier--Stokes solutions are computed using a Fourier spectral solver.

\textbf{Diffusion equation.}
We first consider the diffusion equation, which models a purely dissipative process in which spatial gradients of a scalar field are smoothed by diffusion.
Let \( u(x,y,t) \) denote a scalar quantity defined on the spatial domain
\(
\Omega = (0,1)\times(0,1)
\)
for \( t \ge 0 \).
The governing equation reads
\begin{equation}
\label{eq:Diffusion}
\partial_t u = \nu \Delta u
\quad \text{in } \Omega \times (0,T],
\end{equation}
where \( \nu > 0 \) denotes the diffusion coefficient and
\( \Delta = \partial_{xx} + \partial_{yy} \) is the Laplacian operator.
We assume that \(u\) is sufficiently smooth,
for example \(u\in C^{2,1}(\Omega\times(0,T])\),
so that the derivatives in
Eq.~\eqref{eq:Diffusion} are well defined.
The system is equipped with the initial condition
\begin{equation}
\label{eq:Diffusion_IC}
u(x,y,0)
=
\exp\!\left(
-\frac{(x-x_c)^2+(y-y_c)^2}{w_0}
\right)
\sin(\pi x)\sin(\pi y),
\quad (x,y)\in\Omega,
\end{equation}
where the centroid and width parameters $(x_c, y_c, w_0)$ are randomly sampled according to the uniform distributions specified in Extended Data Table~\ref{table:IC-Parameters}. 
The sinusoidal taper ensures smooth decay toward the boundary.
On the boundary \( \partial\Omega \), we impose the homogeneous
Neumann boundary condition
\begin{equation}
\nabla u\cdot\mathbf{n}=0
\quad \text{on } \partial\Omega\times(0,T],
\end{equation}
where \( \mathbf{n} \) denotes the outward unit normal vector.

\textbf{Advection equation.}
We next consider the advection equation, which describes the transport of a scalar field by a prescribed uniform velocity field.
The governing equation is given by
\begin{equation}
\label{eq:Advection}
\partial_t u + \mathbf{a}\cdot\nabla u = 0
\quad \text{in } \Omega \times (0,T],
\end{equation}
where \( \mathbf{a} = (a_x, a_y) \in \mathbb{R}^2 \) is a constant advection velocity vector.
We assume sufficient regularity of the solution, e.g.,
\(u \in C^{1}(\Omega \times (0,T])\), so that the derivatives in
Eq.~\eqref{eq:Advection} exist.
The advection problem is posed with the same initial condition \eqref{eq:Diffusion_IC} and the homogeneous Neumann boundary condition
\begin{equation}
\label{eq:Advection_BC}
\nabla u \cdot \mathbf{n} = 0
\quad \text{on } \partial\Omega \times (0,T].
\end{equation}
The solution corresponds to a translation of the initial profile along the flow direction prescribed by \( \mathbf{a} \),
while preserving the overall shape and amplitude.

\textbf{Advection--diffusion equation.}
We consider the advection--diffusion equation, which describes
the combined effects of advective transport and diffusive spreading.
The governing equation takes the form
\begin{equation}
\partial_t u + \mathbf{a}\cdot\nabla u
= \nu \Delta u
\quad \text{in } \Omega\times(0,T],
\label{eq:AdvDiff}
\end{equation}
where \( \mathbf{a} = (a_x, a_y) \in \mathbb{R}^2 \) is a constant advection velocity
and \( \nu>0 \) denotes the diffusion coefficient.
A sufficiently smooth solution is assumed,
for instance \(u\in C^{2,1}(\Omega\times(0,T])\),
ensuring that Eq.~\eqref{eq:AdvDiff} is properly defined.
The system is supplemented by the initial condition
\eqref{eq:Diffusion_IC} and a homogeneous Neumann boundary condition
\begin{equation}
\nabla u\cdot\mathbf n=0
\quad \text{on } \partial\Omega\times(0,T].
\end{equation}
The solution exhibits advective transport together with diffusive smoothing,
resulting in spreading and amplitude decay over time.

\textbf{Advection--diffusion--reaction equation.}
We consider the advection--diffusion--reaction equation, which describes
the combined effects of advective transport, diffusive spreading, and local reaction.
The governing equation takes the form
\begin{equation}
\partial_t u + \mathbf{a}\cdot\nabla u
= \nu \Delta u + R(u)
\quad \text{in } \Omega\times(0,T],
\label{eq:AdvDiffReact}
\end{equation}
where \( \mathbf{a}\in\mathbb{R}^2 \) is a constant advection velocity,
\( \nu>0 \) denotes the diffusion coefficient, and \(R(u)\) is a reaction term.
A sufficiently smooth solution is assumed,
for instance \(u\in C^{2,1}(\Omega\times(0,T])\),
ensuring that Eq.~\eqref{eq:AdvDiffReact} is properly defined.
The system is supplemented by the initial condition
\eqref{eq:Diffusion_IC} and a homogeneous Neumann boundary condition
\begin{equation}
\nabla u\cdot\mathbf n=0
\quad \text{on } \partial\Omega\times(0,T].
\end{equation}
In this study, we consider the linear reaction term
\( R(u) = k u \), with \( k \ge 0 \), representing local growth dynamics.
When \( R(u) = 0 \), the equation reduces to the advection--diffusion equation.

\textbf{Allen--Cahn equation.}
We consider the Allen--Cahn equation, a prototypical nonlinear reaction--diffusion model that describes phase separation and interface dynamics in bistable systems.
Let \( u(x,y,t) \) denote an order parameter defined on the spatial domain
\(
\Omega = (0,1)\times(0,1)
\)
for \( t \ge 0 \).
The governing equation is given by
\begin{equation}
\label{eq:AllenCahn}
\partial_t u
=
\varepsilon^{2}\Delta u
-
\frac{1}{\varepsilon^{2}}\bigl(u^{3}-u\bigr)
\quad \text{in } \Omega \times (0,T],
\end{equation}
where \( \varepsilon>0 \) is a small parameter controlling the interfacial thickness and
\( \Delta = \partial_{xx} + \partial_{yy} \) denotes the Laplacian operator.
We restrict attention to sufficiently smooth solutions \(u\),
e.g., \(u\in C^{2,1}(\Omega\times(0,T])\).
The system is initialized with the same initial condition
defined in Eq.~\eqref{eq:Diffusion_IC}.
On the boundary of the domain, we impose a homogeneous Dirichlet boundary condition,
\begin{equation}
\label{eq:AllenCahn_BC}
u = 0
\quad \text{on } \partial\Omega \times (0,T].
\end{equation}
This condition fixes the phase variable at the boundary and prevents interface motion across the domain boundary.
Over time, the solution reflects the combined effects
of diffusion and nonlinear reaction,
leading to smoothing and phase separation behavior.

\textbf{Burgers' equation.}
The two-dimensional Burgers' equation is a nonlinear model
combining convective transport and viscous diffusion.
The equation describes the evolution of a velocity field
$
\mathbf{u}(x,y,t)
=
\left(
u_1(x,y,t), \,
u_2(x,y,t)
\right)
$
defined on the square domain
$ \Omega = (-1,1)\times(-1,1) $,
and takes the form
\begin{equation}
\label{eq:Burgers_vec}
\partial_t \mathbf{u}
+ (\mathbf{u}\cdot\nabla)\,\mathbf{u}
= \nu\,\Delta \mathbf{u},
\end{equation}
where \( \nu>0 \) denotes the kinematic viscosity.
We consider sufficiently smooth velocity fields,
for example
\(\mathbf{u}\in C^{2,1}(\Omega\times(0,T])\),
so that all derivatives in Eq.~\eqref{eq:Burgers_vec}
are well defined.
Written in component form, the system becomes
\begin{equation}
\label{eq:Burgers_comp}
\begin{aligned}
\frac{\partial u_1}{\partial t}
+ u_1 \frac{\partial u_1}{\partial x}
+ u_2 \frac{\partial u_1}{\partial y}
&= \nu \left(
\frac{\partial^2 u_1}{\partial x^2}
+ \frac{\partial^2 u_1}{\partial y^2}
\right), \\
\frac{\partial u_2}{\partial t}
+ u_1 \frac{\partial u_2}{\partial x}
+ u_2 \frac{\partial u_2}{\partial y}
&= \nu \left(
\frac{\partial^2 u_2}{\partial x^2}
+ \frac{\partial^2 u_2}{\partial y^2}
\right).
\end{aligned}
\end{equation}
The initial condition consists of spatially localized velocity fields
with Gaussian profiles, modulated by a sine taper
to satisfy homogeneous Dirichlet boundary conditions.
Specifically,
\begin{equation}
\label{eq:Burgers_IC}
\begin{aligned}
u_1(x,y,0)
&=
\exp\!\left[
-\frac{(x-c_{x,1})^2 + (y-c_{y,1})^2}{w_1}
\right]
\sin(\pi x)\sin(\pi y), \\
u_2(x,y,0)
&=
\exp\!\left[
-\frac{(x-c_{x,2})^2 + (y-c_{y,2})^2}{w_2}
\right]
\sin(\pi x)\sin(\pi y),
\end{aligned}
\end{equation}
where the component centroids $(c_{x,i}, c_{y,i})$ and width parameters $w_i$ (for $i=1,2$) are randomly sampled according to the uniform distributions defined in Extended Data Table~\ref{table:IC-Parameters}.
The velocity field satisfies homogeneous Dirichlet boundary conditions,
\begin{equation}
\mathbf{u}(x,y,t)=\mathbf{0},
\qquad (x,y)\in\partial\Omega.
\end{equation}
The nonlinear term \((\mathbf{u}\cdot\nabla)\mathbf{u}\)
describes advective transport of the velocity field,
while the viscous term \(\nu\Delta\mathbf{u}\)
introduces diffusive smoothing.
The solution behavior is determined by the relative magnitude
of convection and viscosity.

\textbf{Incompressible Navier--Stokes equations.}
We next consider the two-dimensional incompressible Navier--Stokes equations, which describe the evolution of a divergence-free velocity field driven by nonlinear advection and balanced by pressure and viscous diffusion.
Let \( \mathbf{u}(x,y,t) = (u_{1}(x,y,t),\,u_{2}(x,y,t)) \) denote the velocity field and \( p(x,y,t) \) the kinematic pressure on the spatial domain
\(
\Omega = (0,L)\times(0,L)
\)
for \( t\ge 0 \).
The governing equations without external forcing read
\begin{equation}
\label{eq:NSE2D}
\begin{aligned}
\partial_t \mathbf{u} + (\mathbf{u}\cdot\nabla)\mathbf{u} + \nabla p &= \nu\,\Delta \mathbf{u},
\quad &&\text{in } \Omega\times(0,T],\\
\nabla\cdot\mathbf{u} &= 0,
\quad &&\text{in } \Omega\times[0,T],
\end{aligned}
\end{equation}
where \( \nu>0 \) is the kinematic viscosity.
We assume sufficiently smooth velocity and pressure fields
\((\mathbf{u},p)\) so that Eq.~\eqref{eq:NSE2D}
is well defined.
The system is initialized by a solenoidal velocity field parameterized by an amplitude factor $a$, sampled from the uniform distribution specified in Extended Data Table~\ref{table:IC-Parameters}. For all \( (x,y)\in\Omega \), the initial condition is defined as
\begin{equation}
\label{eq:NSE2D_IC}
\begin{aligned}
u_{1}(x,y,0) &= -a\,\phi\!\Bigl(\frac{2 \pi y}{L}\Bigr),\\
u_{2}(x,y,0) &= \ \ a\,\psi\!\Bigl(\frac{4\pi x}{L}\Bigr),
\end{aligned}
\end{equation}
where each of \( \phi \) and \( \psi \) is independently chosen
from \( \{\sin(\cdot),\cos(\cdot)\} \), resulting in four possible
sine–cosine combinations.
This construction satisfies the incompressibility constraint at \(t=0\),
since \(u_1\) depends only on \(y\) and \(u_2\) only on \(x\),
which directly implies
\(
\nabla\!\cdot\mathbf{u}(\cdot,\cdot,0)=0.
\)
We impose periodic boundary conditions for both velocity and pressure,
\begin{equation}
\label{eq:NSE2D_BC}
\begin{aligned}
\mathbf{u}(x+L,y,t) &= \mathbf{u}(x,y,t),&
\mathbf{u}(x,y+L,t) &= \mathbf{u}(x,y,t),\\
p(x+L,y,t) &= p(x,y,t),&
p(x,y+L,t) &= p(x,y,t),
\end{aligned}
\end{equation}
for all \( (x,y)\in\Omega \) and \( t\in[0,T] \).

\subsubsection*{Evaluation metrics}\label{sec:metric}

We employ a set of evaluation metrics to assess accuracy, robustness to physical misspecification, and statistical reliability.

\textbf{Pointwise accuracy.} Pointwise accuracy is quantified using the relative $L_2$ percentage error:
\begin{equation}
\mathrm{Rel}\text{-}L_2 (\%) = 100 \times \frac{\| u_{\text{pred}} - u_{\text{true}} \|_2}{\| u_{\text{true}} \|_2}.
\end{equation}
As pADAM is a generative model, point predictions are obtained by drawing a single sample from the learned solution distribution for each test case. Reported errors are averaged over 50 independent test instances for each problem setting.

\textbf{Quantifying operator shift.} To assess the generative prior's robustness to out-of-distribution (OOD) physical dynamics, we define the operator shift $\Delta_{\mathrm{op}}$ as a measure of the discrepancy between the trained physical library and an unseen target dynamic. For a target operator $\mathcal{P}_{\text{unseen}}$ and a reference training operator $\mathcal{P}_{\text{train}}$, the shift is quantified by the relative $L_2$ deviation of their respective terminal states $u_T$ evolving from an identical initial condition $u_0$:
\begin{equation}
\label{eq:op_shift}
\Delta_{\mathrm{op}}(\%) = 100 \times \frac{\|u_T^{(\mathcal{P}_{\text{unseen}})} - u_T^{(\mathcal{P}_{\text{train}})}\|_2}{\|u_T^{(\mathcal{P}_{\text{train}})}\|_2}.
\end{equation}
In the context of the zero-shot extrapolation experiments presented in this study, this metric captures the physical departure induced by the reaction term $k$ in the advection--diffusion--reaction (ADR) system relative to the base advection--diffusion (AD) trajectory. This serves as a formal proxy for the degree of physical misspecification the pADAM prior must reconcile during observation-guided sampling.

\textbf{Uncertainty quantification.} The reliability of uncertainty quantification is evaluated using the prediction interval coverage probability (PICP):
\begin{equation}
\mathrm{PICP} (\%) = \frac{100}{|\mathcal{S}|} \sum_{\mathbf{s} \in \mathcal{S}} \mathds{1}\bigl\{\, u_{\text{true}}(\mathbf{s}) \in [\,\hat{u}_{\mathrm{low}}(\mathbf{s}),\, \hat{u}_{\mathrm{high}}(\mathbf{s})\,] \bigr\},
\end{equation}
where $\mathcal{S}$ denotes the set of spatial coordinates and $\mathds{1}\{\cdot\}$ represents the indicator function.

\section*{Data availability}

The data that support the findings of this study are available from the corresponding author upon reasonable request.

\section*{Code availability}

The code used to generate the results of this study will be made publicly available upon publication at the following repository: \url{https://github.com/Mollaali/pADAM}.

\section*{Acknowledgement}
We would like to thank the support of National Science Foundation (DMS-2533878, DMS-2053746, DMS-2134209, ECCS-2328241, CBET-2347401 and OAC-2311848), and U.S.~Department of Energy (DOE) Office of Science Advanced Scientific Computing Research program DE-SC0023161, the SciDAC LEADS Institute, and DOE–Fusion Energy Science, under grant number: DE-SC0024583.
\backmatter
\bibliography{bibliography}

\clearpage
\section*{Extended Data}

\setcounter{figure}{0}
\renewcommand{\figurename}{Extended Data Fig.}

\setcounter{table}{0}
\renewcommand{\tablename}{Extended Data Table}

\begin{table}[t!]
\caption{
\textbf{Experimental configurations and parameter manifolds.}
\textbf{a,} Initial condition (IC) protocol standardized across all experimental regimes. ICs for scalar and Burgers' systems follow localized Gaussian forms adapted to their respective domains (Eqs.~\ref{eq:Diffusion_IC}, \ref{eq:Burgers_IC}), while Navier--Stokes follows the solenoidal trigonometric form (Eq.~\ref{eq:NSE2D_IC}).
\textbf{b,} Systematic variability of governing coefficients $\phi$ across thematic investigations. These manifolds define the continuous physical regimes analyzed in the corresponding Results sections.}
\label{table:ed_combined}
\centering

\renewcommand{\arraystretch}{1.2}
\small

\begin{tabular*}{\linewidth}{@{\extracolsep{\fill}}p{3.4cm} p{2.1cm} p{2.5cm} l@{}}
\textbf{PDE group} & \multicolumn{2}{l}{\textbf{IC parameter}} & \textbf{Sampling rule} \\
\midrule
\multirow[t]{2}{*}{Scalar PDEs}      & \multicolumn{2}{l}{Centroid $(x_c, y_c)$}          & $\mathcal{U}[0.2, 0.8]^2$ \\
                                     & \multicolumn{2}{l}{Gaussian Width $w_0$}           & $\mathcal{U}[0.025, 0.075]$ \\
                                     
\midrule
\multirow[t]{2}{*}{Burgers'} & \multicolumn{2}{l}{Component Centroids $(c_{x,i}, c_{y,i})$} & $\mathcal{U}[0.2, 0.8]^2$ \\
                                     & \multicolumn{2}{l}{Component Widths $w_1, w_2$}    & $\mathcal{U}[0.025, 0.075]$ \\

\midrule
\multirow[t]{4}{*}{Navier--Stokes}
  & \multicolumn{2}{l}{Amplitude Factor $a$}            & $a \sim \mathcal{U}(0.5, 1.5]$ \\
  & \multicolumn{2}{l}{Trigonometric Basis $\phi, \psi$}& $(\phi,\psi) \in \{\sin,\cos\}^2$ \\
  & \multicolumn{2}{l}{Mode Frequency}                   & $k \in \{2\pi, 4\pi\}$ \\
  & \multicolumn{2}{l}{Domain Length $L$}                & $L = 1$ (fixed) \\

\midrule
\addlinespace[12pt]
\multicolumn{4}{@{}p{\linewidth}@{}}{\centering\textbf{Panel b | Parameter manifolds and physical regimes across thematic investigations}} \\
\midrule
\textbf{Thematic investigation} & \textbf{PDE library} & \textbf{Variable parameters ($\phi$)} & \textbf{Sampling rule} \\
\midrule
\multirow[t]{3}{3.4cm}{\textbf{Unified multi-physics} }  & Diffusion & -- & $\nu = 0.25$ \\
                                           & Advection & -- & $(a_x, a_y) = (4, 2)$ \\
                                           & Adv.--Diff. & -- & $(\nu, a_x, a_y) = (0.25, 4, 2)$ \\
\midrule
\multirow[t]{3}{3.4cm}{\textbf{Continuous physics manifold} }   & Diffusion & $\nu$ & $\mathcal{U}[0.1, 0.4]$ \\
                                           & Advection & $a_x$ & $\mathcal{U}[2.0, 5.0], a_y=2$ \\
                                           & Adv.--Diff. & $\nu$ & $\mathcal{U}[0.1, 0.4], (a_x, a_y)=(4, 2)$ \\
\midrule
\multirow[t]{6}{3.4cm}{\textbf{Structural scaling} }     & Diffusion & $\nu$ & $\mathcal{U}[0.1, 0.4]$ \\
                                           & Advection & $a_x$ & $\mathcal{U}[2.0, 5.0], a_y=2$ \\
                                           & Adv.--Diff. & $\nu$ & $\mathcal{U}[0.1, 0.4], (a_x, a_y)=(4, 2)$ \\
                                           & Allen--Cahn & $\varepsilon^2$ & $\mathcal{U}[2.5 \times {10^{-3}}, 0.0121]$ \\
                                           & Burgers' & $\nu$ & $\nu = 0.05$ \\
                                           & Navier--Stokes & $\nu$ & $\nu = 0.02$ \\
\midrule
\multirow[t]{3}{3.4cm}{\textbf{Parametric scaling and model selection}} & Diffusion & $\nu$ & $\mathcal{U}[0.2, 0.4]$ \\
                                           & Advection & $(a_x, a_y)$ & $\mathcal{U}[2.0, 3.0]^2$ \\
                                           & Adv.--Diff. & $(\nu, a_x, a_y)$ & $\mathcal{U}[0.2, 0.4] \times \mathcal{U}[2.0, 3.0]^2$ \\
\bottomrule
\end{tabular*}
\label{table:IC-Parameters}
\end{table}

\begin{figure}[ht]
    \centering
    \includegraphics[width=0.95\linewidth]{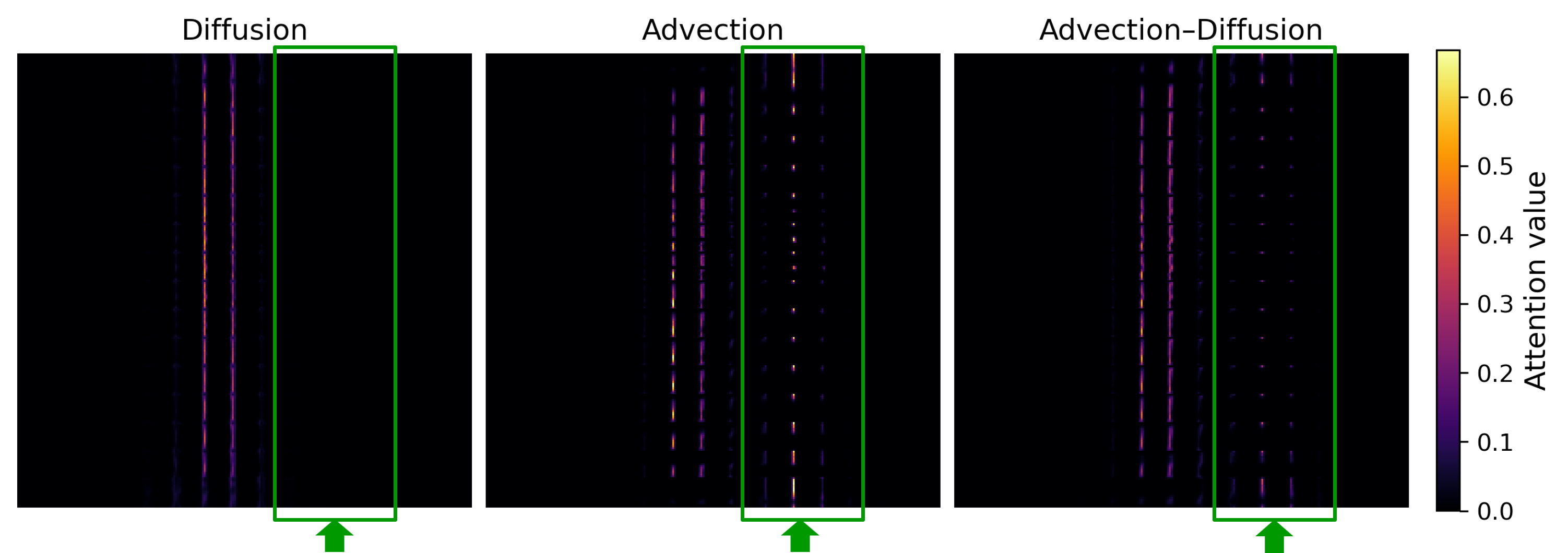}
    \caption{\textbf{Mechanistic analysis of internal attention patterns across physical operators.} 
    Representative attention maps are extracted from a decoder block at a fixed denoising step ($t=1000$) for diffusion, advection, and advection--diffusion regimes under identical initial conditions. 
    Comparison of local attention weights reveals distinct structural signatures: the advection--diffusion maps (right) exhibit intermediate activations in spatial regions where features are prominent in pure advection (middle) but absent in pure diffusion (left). This graded attention pattern across the three operator classes suggests that the shared pADAM prior reuses and composes operator-specific internal representations to capture complex mixed dynamics.
     }
    \label{fig:exp1:attention_mechanism}
\end{figure}

\begin{figure}[ht]
    \centering
    \includegraphics[width=\linewidth]{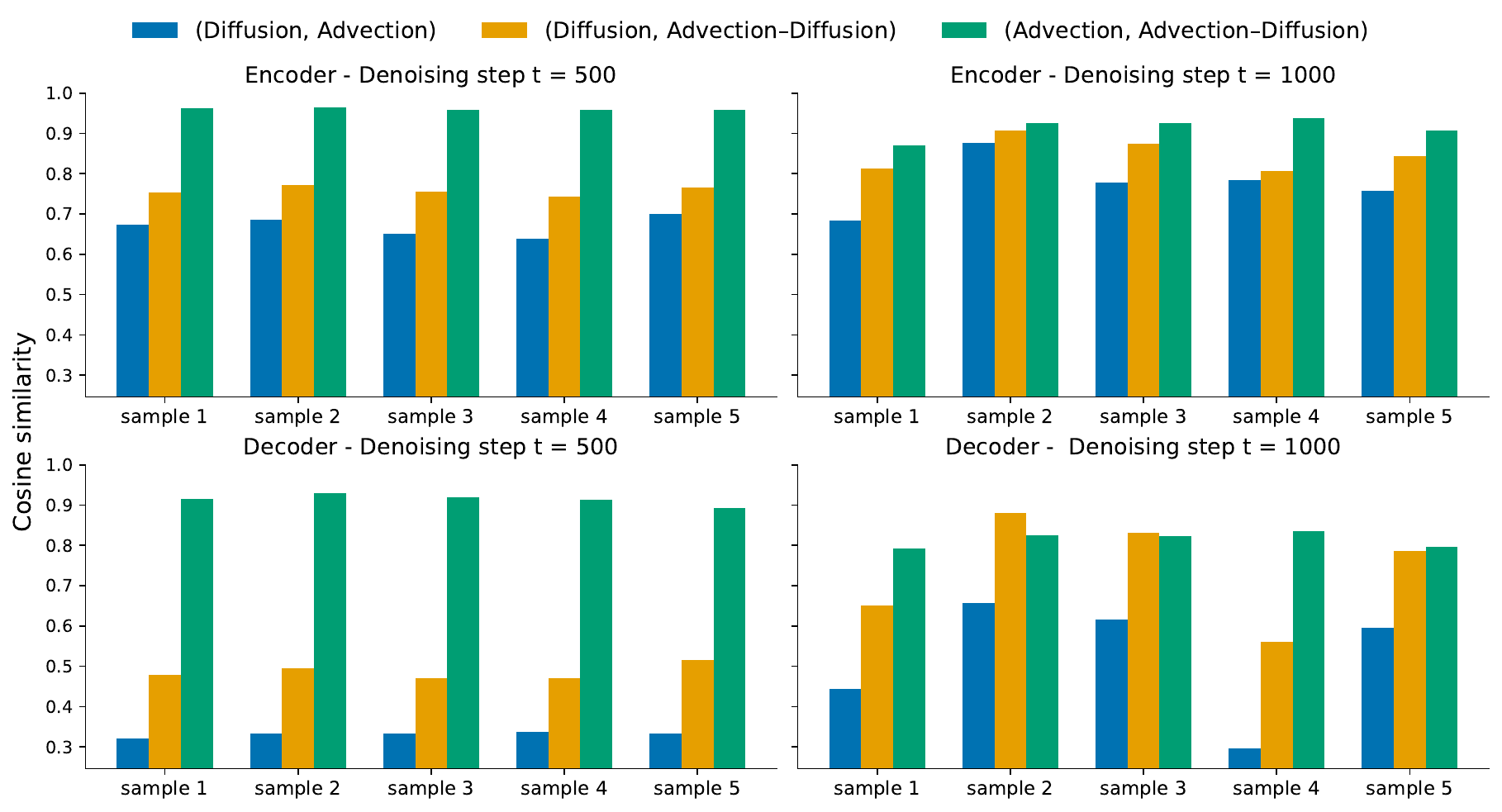} 
    \caption{\textbf{Quantitative assessment of representational similarity across PDE families.} 
    Pairwise cosine similarities of attention maps calculated for one encoder and one decoder block at two representative denoising steps ($t=500$ and $t=1000$) across five test samples. 
    Across all configurations, similarities between the advection--diffusion (mixed) regime and the pure extremes (diffusion or advection) are consistently higher than the similarity between the two pure extremes. This hierarchy indicates that the model's latent space is organized according to the underlying mathematical composition of the physical laws, with the mixed operator acting as a representational bridge.
    }
    \label{fig:exp1:attention_similarity}
\end{figure}

\begin{figure}[ht]
\centering
\begin{subfigure}[t]{0.38\linewidth}
\centering
\caption{Operator discrepancy ($\Delta_{\mathrm{op}}$)}
\label{fig:op_shift_table}
\renewcommand{\arraystretch}{1.1}
\begin{tabular}{cS}
\toprule
Reaction rate $k$ & {$\Delta_{\mathrm{op}}$ (\%)} \\
\midrule
5.0 & 22.14 \\
15.0 & 52.83 \\
\bottomrule
\end{tabular}
\end{subfigure}
\hfill
\begin{subfigure}[t]{0.58\linewidth}
\centering
\caption{ADR extrapolation performance}
\label{fig:adr_extrap_plot}
\includegraphics[width=\linewidth]{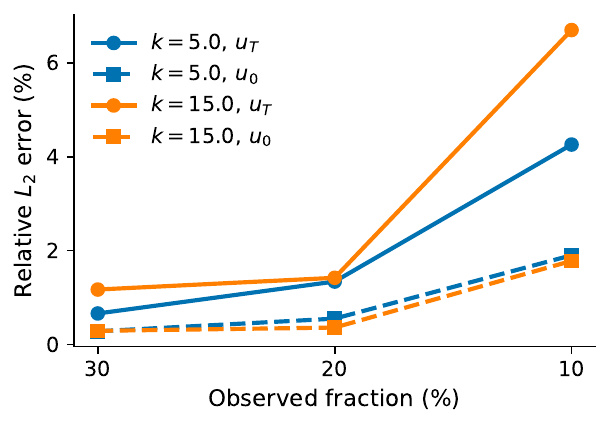}
\end{subfigure}

    \caption{\textbf{Zero-shot extrapolation performance on unseen PDE (advection--diffusion--reaction dynamics).} 
    \textbf{a,} Quantification of the operator shift ($\Delta_{\mathrm{op}}$) between the advection--diffusion (AD) training prior and the unseen advection--diffusion--reaction (ADR) dynamics for two reaction rates $k$. This shift is defined as the relative $L_2$ discrepancy between the terminal states ($u_T$) of the two systems under identical initial conditions; as $k$ increases, the physical divergence between the two PDE systems grows.
    \textbf{b,} Relative $L_2$ error for the joint reconstruction of full-field initial ($u_0$) and terminal ($u_T$) states conditioned on sparse spatial observations of endpoints. The model is conditioned on the closest known operator class (AD) and steered via observation-guided sampling. Error in $u_T$ increases with $k$ due to the accumulated influence of the unseen reaction term, while $u_0$ remains stable. As observation sparsity increases, reconstruction accuracy degrades gracefully, demonstrating the robustness of the pADAM prior under physical misspecification. All results are averaged over 20 independent test instances.
    }
\label{fig:adr_extrap_combined}
\end{figure}

\begin{figure}[ht]
    \centering
    \includegraphics[width=\linewidth]{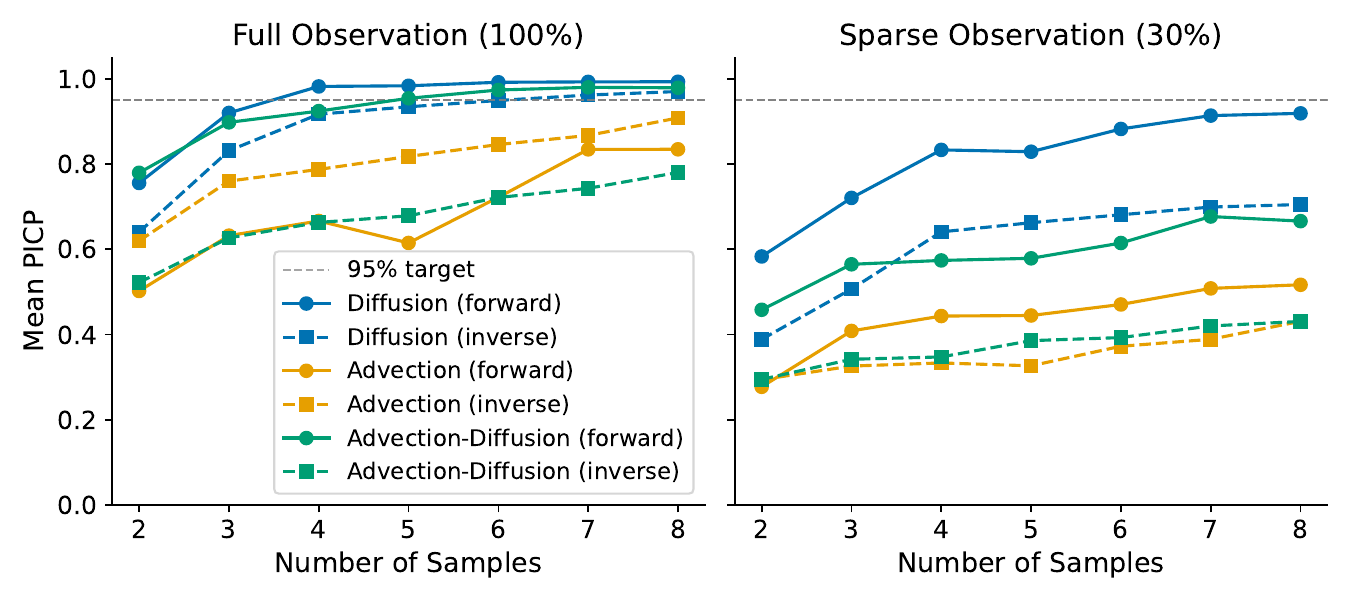}
    \caption{\textbf{Empirical coverage of Bayesian posterior ensembles across different observation regimes.} 
    Mean predictive-interval coverage probability (PICP) of nominal 95\% intervals (dashed grey line) as a function of ensemble size for forward and inverse tasks. 
    Particularly under ill-posed conditions, such as inverse problems or prediction under sparse observations, where only 30\% of the field is observed, the coverage of raw Bayesian intervals saturates well below the 95\% target across all operator families. This systematic under-coverage underscores the necessity of conformal calibration for providing rigorous reliability guarantees in data-sparse or ill-posed regimes. Results are averaged over 20 test instances.
    }
    \label{fig:exp1:PICP_vs_samples}
\end{figure}

\begin{table}[ht]
    \centering
    \caption{\textbf{Conformal calibration resolves the systematic under-coverage of Bayesian ensembles.} 
    Mean empirical coverage (\%) of nominal 95\% prediction intervals for the advection--diffusion system under sparse (30\%) spatial observations (ensemble size $M=6$; 50 calibration instances). While raw Bayesian ensembles fail to meet the nominal target, the integration of conformal calibration into the pADAM framework effectively recovers statistical validity, ensuring reliable uncertainty quantification for physical inference. Results are averaged over 50 test instances.
    }
    \vspace{2pt}
    \begin{minipage}{0.75\linewidth}
        \centering
        \small
        \renewcommand{\arraystretch}{1.3}
        \begin{tabular*}{\linewidth}{@{\extracolsep{\fill}} l cc @{}}
        \toprule
        \textbf{Method} & \textbf{Forward ($u_T$)} & \textbf{Inverse ($u_0$)} \\
        \midrule
        Ensemble only             & 58.33 & 36.31 \\
        Ensemble + Conformal      & \textbf{98.42} & \textbf{99.83} \\
        \bottomrule
        \end{tabular*}
    \end{minipage}
    \label{table:exp1_conformal}
\end{table}

\begin{figure}[ht]
    \centering
    \includegraphics[width=1\linewidth]{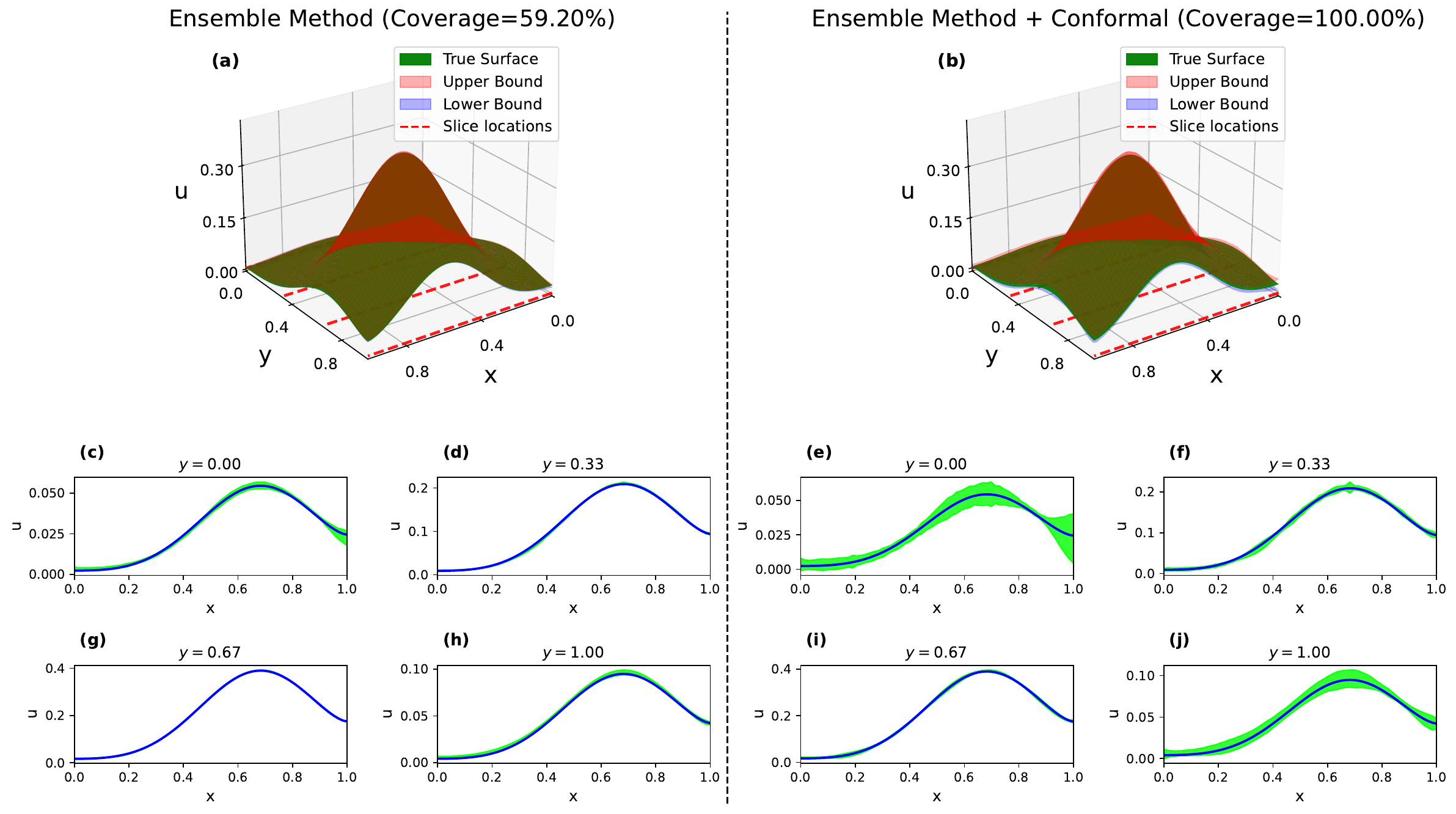}
    \caption{\textbf{Effect of conformal calibration on prediction intervals.} 
    \textbf{a,} Standard ensemble-based prediction intervals show undercoverage (59.20\%) for forward prediction of $u_T$ in the advection--diffusion system under sparse (30\%) observations.
    \textbf{b,} Conformal-calibrated intervals achieving 100.00\% empirical coverage. 
    \textbf{c--j,} One-dimensional slices at representative $y$-locations; shaded bands denote predictive intervals and blue curves indicate the true solution. Conformal calibration adaptively expands the intervals in regions of high epistemic uncertainty to recover the nominal 95\% coverage. Critically, this post-hoc procedure preserves the shape of the underlying generative predictions while improving coverage reliability without model retraining.
     }
    \label{fig:exp1:conformal_effect}
\end{figure}

\begin{table}[ht]
\centering
\small
\caption{
    \textbf{Task-agnostic performance across the continuous physics manifold.} Relative $L_2$ errors (\%) for forward prediction ($u_T$), initial-state reconstruction ($u_0$), and physical parameter discovery ($\phi$). pADAM was trained across three PDE families, each with a single variable coefficient: diffusion ($\phi=\nu$), advection ($\phi=a_x$), and advection--diffusion ($\phi=\nu$). Performance is evaluated under full (100\%) and sparse spatial observations (30\% and 10\%), with results averaged over 50 test instances.}
\renewcommand{\arraystretch}{1.2}
\small 
\begin{tabular}{llccc}
\toprule
\textbf{PDE System} & \textbf{Observation} & \textbf{Forward ($u_T$)} & \textbf{Inverse ($u_0$)} & \textbf{Inverse ($\phi$)} \\
\midrule
\multirow{3}{*}{Diffusion}  
 & Full (100\%)    & 0.69 & 0.89 & 1.38 \\
 & Sparse (30\%)   & 1.64 & 1.96 & 3.48 \\
 & Sparse (10\%)   & 2.09 & 3.05 & 8.26 \\
\midrule
\multirow{3}{*}{Advection}  
 & Full (100\%)    & 1.91 & 1.13 & 0.72 \\
 & Sparse (30\%)   & 2.11 & 1.47 & 1.48 \\
 & Sparse (10\%)   & 2.13 & 2.55 & 2.69 \\
\midrule
\multirow{3}{*}{Advection--diffusion}  
 & Full (100\%)    & 1.12 & 1.20 & 2.81 \\
 & Sparse (30\%)   & 1.70 & 2.26 & 4.44 \\
 & Sparse (10\%)   & 2.65 & 4.11 & 7.73 \\
\bottomrule
\end{tabular}
\label{ext_table:vp}
\end{table}

\begin{figure}[t!]
\centering

\begin{subfigure}{\linewidth}
    \centering
    \includegraphics[width=\linewidth]{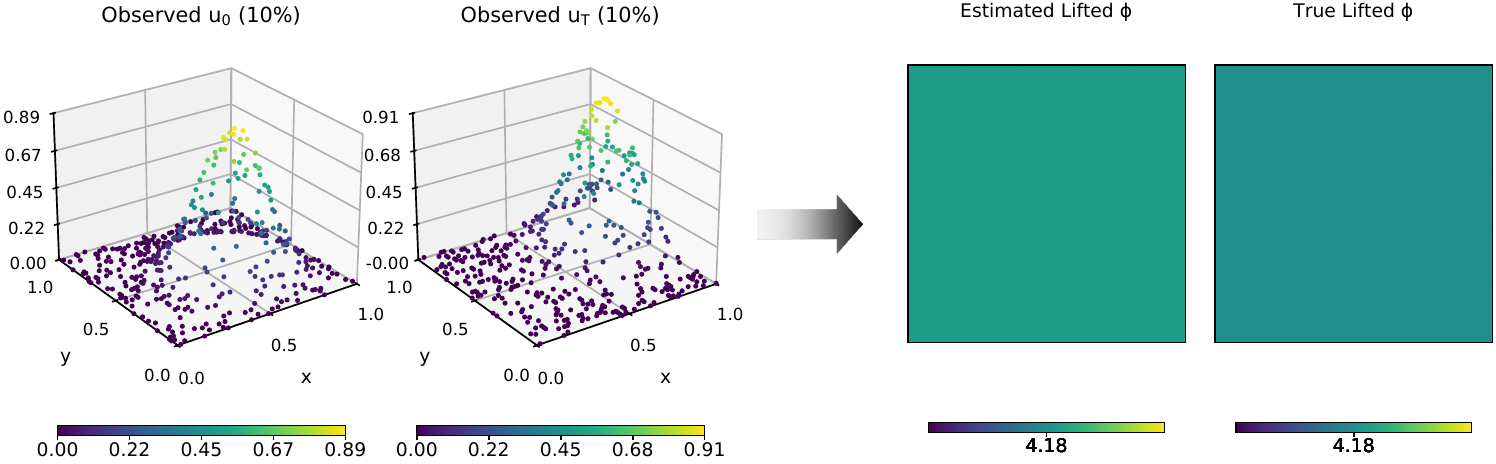}
    \subcaption{}
    \label{fig:manifold_a}
\end{subfigure}

\vspace{6pt}

\begin{subfigure}{\linewidth}
    \centering
    \includegraphics[width=\linewidth]{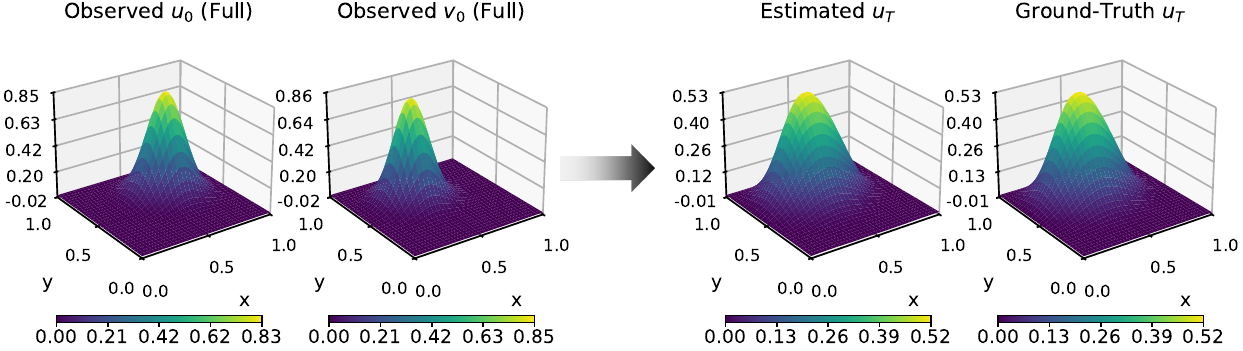}
    \subcaption{}
    \label{fig:burgers_b}
\end{subfigure}

\vspace{6pt}

\begin{subfigure}{\linewidth}
    \centering
    \includegraphics[width=\linewidth]{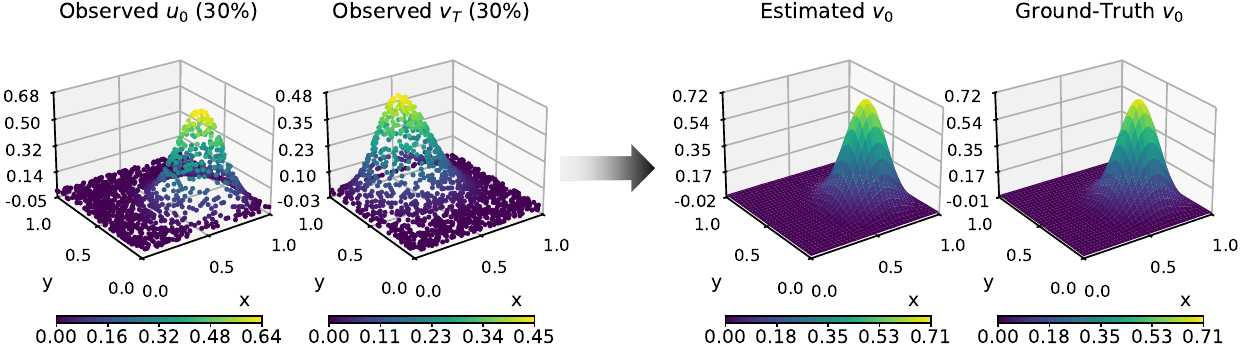}
    \subcaption{}
    \label{fig:burgers_c}
\end{subfigure}

\caption{\textbf{Qualitative inference performance of pADAM under continuous physics manifold and structural scaling.} 
\textbf{a,} Parameter discovery within the scalar physics manifold (diffusion, advection, and advection--diffusion). For the advection system, the physical coefficient $\phi = a_x$ is inferred from paired states $(u_0, u_T)$ under 10\% spatial observations. 
\textbf{b, c,} Inference for the Burgers' system under structural scaling to the 6-PDE library.
\textbf{b,} Forward prediction of the Burgers' velocity component $u_T$ conditioned on full spatial observations of the initial states $(u_0, v_0)$. 
\textbf{c,} Inverse reconstruction of the Burgers' initial velocity component $v_0$ conditioned on 30\% spatial observations of the terminal component $v_T$ and the initial component $u_0$. 
}

\label{fig:composite_results}
\end{figure}

\begin{figure}[t!]
\centering

\begin{subfigure}{\linewidth}
    \centering
    \includegraphics[width=\linewidth]{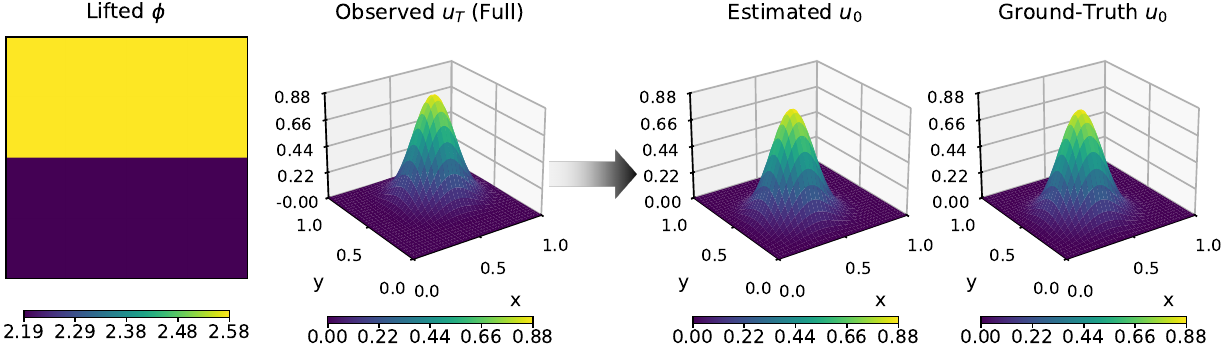}
    \subcaption{}
    \label{fig:hetro_a}
\end{subfigure}

\vspace{6pt}

\begin{subfigure}{\linewidth}
    \centering
    \includegraphics[width=\linewidth]{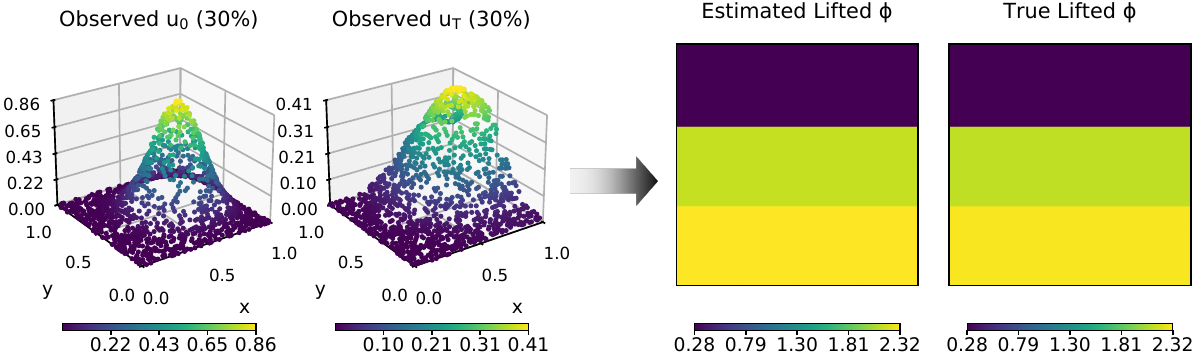}
    \subcaption{}
    \label{fig:hetro_b}
\end{subfigure}

\caption{\textbf{Qualitative inference performance of pADAM under parametric scaling and heterogeneous parametric dimensionality.} 
\textbf{a,} Initial-state reconstruction in the advection system with a variable parameter vector $\phi = [a_x, a_y]$. The reconstructed initial state $u_0$ is inferred by conditioning on the known parameter vector and full observation of the terminal state $u_T$. 
\textbf{b,} Parameter discovery in the advection--diffusion system with a three-dimensional parameter vector $\phi = [\nu, a_x, a_y]$. The physical coefficients are inferred from paired states $(u_0, u_T)$ under 30\% spatial observations.
}
\label{fig:hetro}
\end{figure}


\end{document}